\documentclass[journal]{IEEEtran}
\usepackage{amsfonts}
\usepackage{cite}
\usepackage[cmex10]{amsmath}
\usepackage{color}
\usepackage{flushend}
\usepackage{url}
\usepackage{array}

\usepackage{mdwmath}
\usepackage{mdwtab}
\usepackage{eqparbox}
\usepackage{booktabs} %jia cu

\usepackage{rotating}
\usepackage{afterpage}
\usepackage{placeins}
\usepackage{bbding}
\usepackage{amsmath}   %import split
\usepackage{CJK}    %Chinese in comment
\usepackage[colorlinks]{hyperref}
\usepackage{lineno}
\usepackage{subfigure}   %figure bing lie
\usepackage{graphicx}    %figure bing lie
\newcommand{\myref}[1]{(\ref{#1})}  % define the format of equation reference
\usepackage{multirow}
\usepackage{amssymb}
\usepackage{pifont}% http://ctan.org/pkg/pifont
\usepackage[linesnumbered,boxed,ruled,commentsnumbered]{algorithm2e}%% 算法包，注意设置所需可选项
\usepackage{algpseudocode}
\usepackage{bm}

\usepackage{color}\definecolor{MyDarkGreen}{rgb}{0.8,0.91,0.81}\definecolor{yellow}{rgb}{0.99,0.99,0.70}\definecolor{white}{rgb}{1.0,1.0,1.0}\definecolor{black}{rgb}{0.00,0.00,0.00}
\begin{document}
% \color{MyDarkGreen} zi ti yan se

%\title{R$^3$-Net: A Rotatable Region-Based Residual Network for Multi-Oriented Vehicle Detection in Aerial Images and Videos}
\title{R$^3$-Net: A Deep Network for Multi-oriented Vehicle Detection in Aerial Images and Videos}

\author{Qingpeng~Li,~\IEEEmembership{Student Member,~IEEE,}
        Lichao~Mou,~\IEEEmembership{Student Member,~IEEE,}
        Qizhi~Xu,~\IEEEmembership{Member,~IEEE,}
        Yun~Zhang,~\IEEEmembership{Member,~IEEE,}
        and~Xiao~Xiang~Zhu,~\IEEEmembership{Senior Member,~IEEE}% <-this % stops a space
\thanks{This work was supported in part by the National Natural Science Foundation of China under Grant 61331017 and Grant 61672076, in part by the China Scholarship Council, in part by the European Research Council through the European Union’s Horizon 2020 Research and Innovation Programme (So2Sat) under Grant ERC-2016-StG-714087, in part by the Helmholtz Association under the framework of the Young Investigators Group, SiPEO, under Grant VH-NG-1018, and in part by the Bavarian Academy of Sciences and Humanities in the framework of Junges Kolleg. \emph{(Corresponding author: Qizhi Xu.)}

Q. Li and Q. Xu are with the State Key Laboratory of Virtual Reality Technology and Systems and the Beijing Key Laboratory of Digital Media, School of Computer Science and Engineering, Beihang University, 100191 Beijing, China (e-mail: liqingpeng@buaa.edu.cn; qizhi@buaa.edu.cn).

L. Mou and X. X. Zhu are with the Remote Sensing Technology Institute, German Aerospace Center, 82234 Wessling, Germany and with Signal Processing in Earth Observation, Technical University of Munich, 80333 Munich, Germany (e-mails: lichao.mou@dlr.de; xiao.zhu@dlr.de).

Y. Zhang is with the Canada Research Chair Laboratory in Advanced Geomatics Image Processing, Department of Geodesy and Geomatics Engineering, University of New Brunswick, Fredericton, NB E3B 5A3, Canada (e-mail: yunzhang@unb.ca).
}% <-this % stops a space
\thanks{}}

\markboth{}
{Shell \MakeLowercase{\textit{et al.}}: Bare Advanced Demo of IEEEtran.cls for Journals}
\maketitle

\begin{abstract}
    Vehicle detection is a significant and challenging task in aerial remote sensing applications. Most existing methods detect vehicles with regular rectangle boxes and fail to offer the orientation of vehicles. However, the orientation information is crucial for several practical applications, such as the trajectory and motion estimation of vehicles. In this paper, we propose a novel deep network, called rotatable region-based residual network (R$^3$-Net), to detect multi-oriented vehicles in aerial images and videos. More specially, R$^3$-Net is utilized to generate rotatable rectangular target boxes in a half coordinate system. First, we use a rotatable region proposal network (R-RPN) to generate rotatable region of interests (R-RoIs) from feature maps produced by a deep convolutional neural network. Here, a proposed batch averaging rotatable anchor (BAR anchor) strategy is applied to initialize the shape of vehicle candidates. Next, we propose a rotatable detection network (R-DN) for the final classification and regression of the R-RoIs. In R-DN, a novel rotatable position sensitive pooling (R-PS pooling) is designed to keep the position and orientation information simultaneously while downsampling the feature maps of R-RoIs. In our model, R-RPN and R-DN can be trained jointly. We test our network on two open vehicle detection image datasets, namely DLR 3K Munich Dataset and VEDAI Dataset, demonstrating the high precision and robustness of our method. In addition, further experiments on aerial videos show the good generalization capability of the proposed method and its potential for vehicle tracking in aerial videos. The demo video is available at \url{https://youtu.be/xCYD-tYudN0}.
\end{abstract}


\section{Introduction}% 1.
\label{sec:introduction}

\begin{figure}[tb!]
\centering
\subfigure{
\includegraphics[width=85mm]{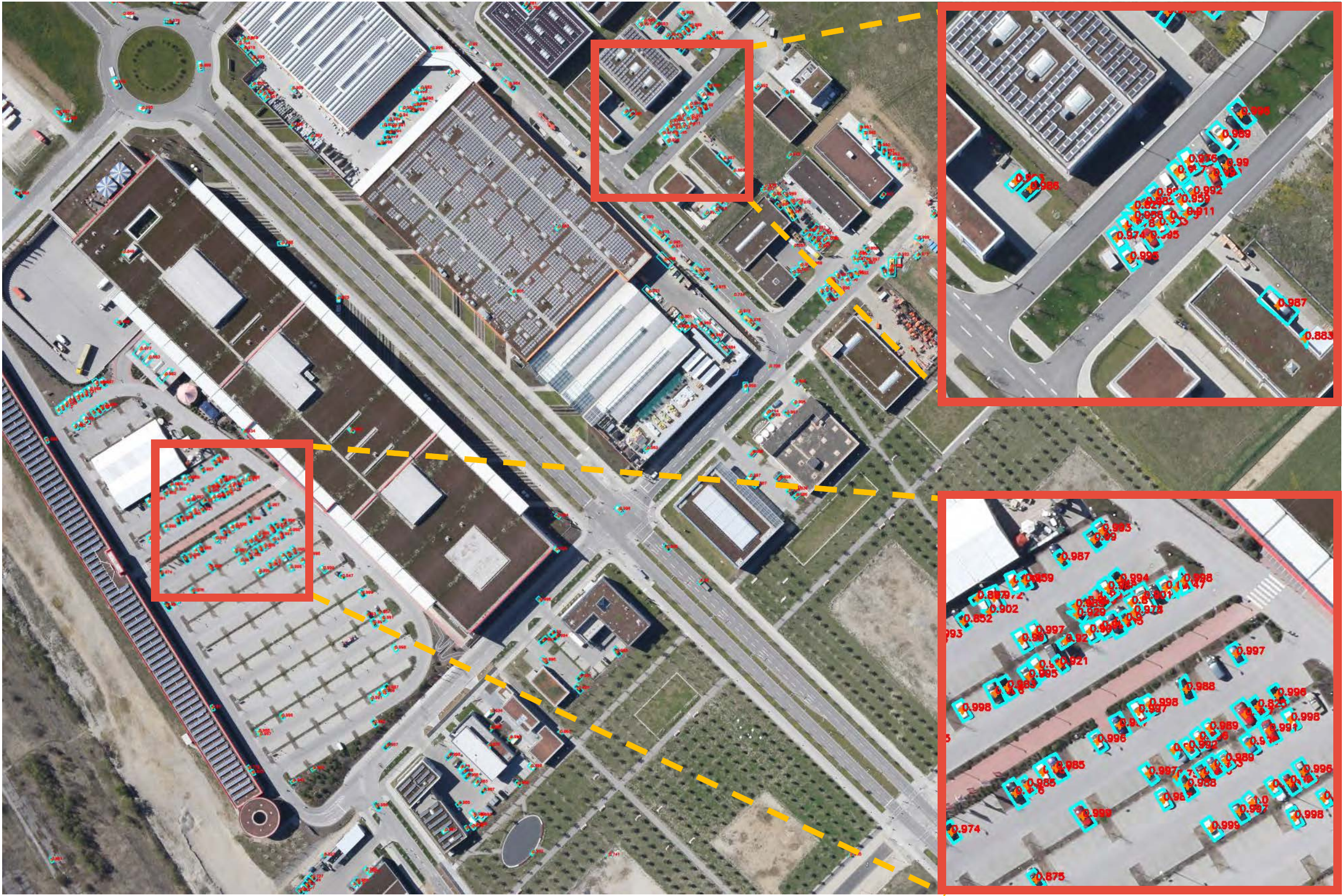}}
\subfigure{
\includegraphics[width=85mm]{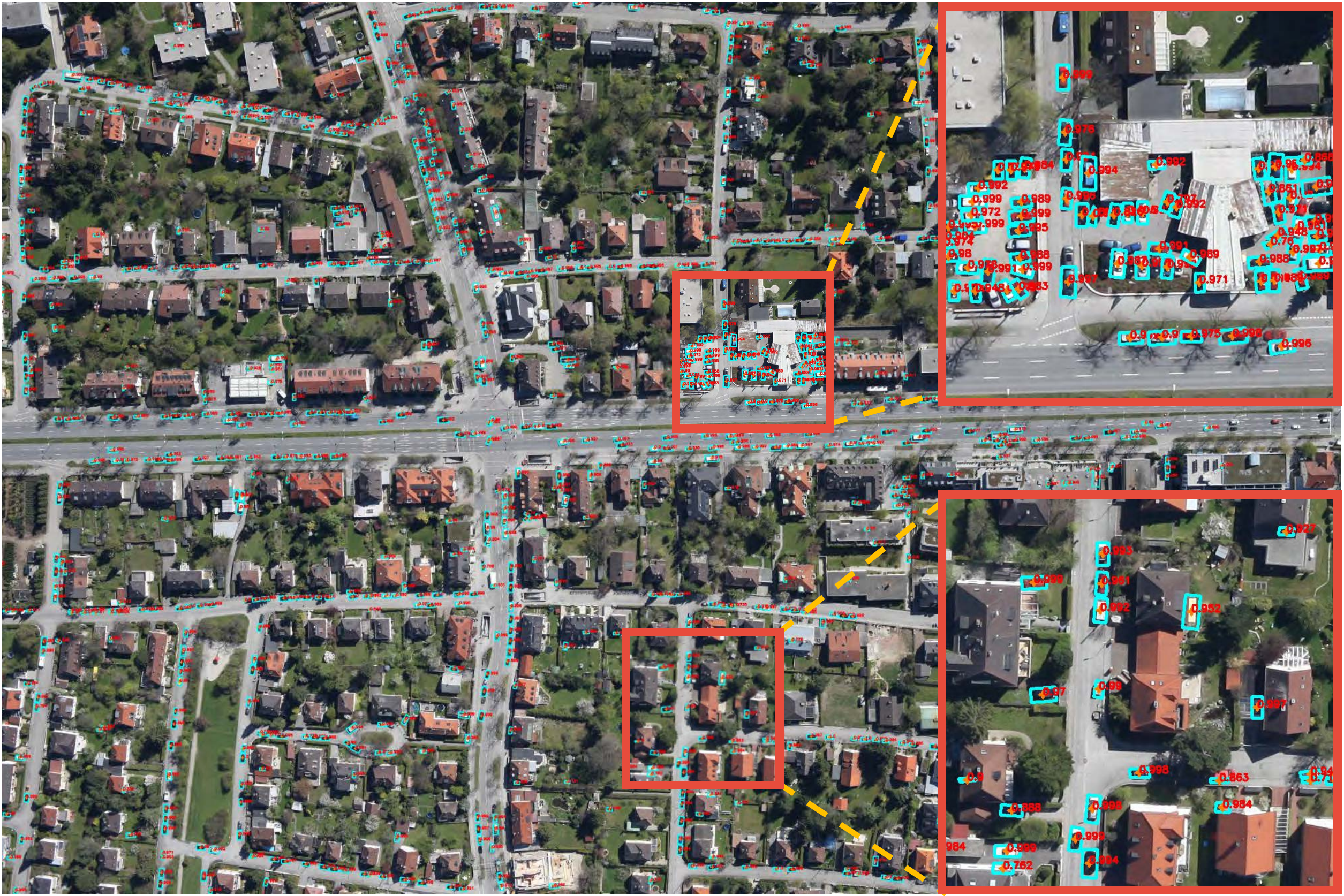}}
\caption{Examples of multi-oriented vehicle detection produced with the proposed network, over two scenes taken from DLR 3K Munich Dataset. Best viewed zoomed in.}
\label{fig:FirstShow} %% label for entire figure
\end{figure}

\IEEEPARstart{A}{LONG} with the now widespread availability of aeroplanes and unmanned aerial vehicles (UAVs), the detection and localization of small targets in high resolution airborne imagery have been attracting a lot of attentions in the remote sensing community~\cite{aerial2017zhangliangpei,UAV2017Tuia,UAV2015Moranduzzo,Audebert2017Beyond,Xia_2018_CVPR,Mou2018RiFCN}. They have numerous useful applications, to name a few, surveillance, defense, and traffic planning~\cite{2016zhangplaneDet,Mou2017tracklet,Mou2017fusion,AudebertRS17,mine2018}. In this paper, vehicles are considered the small targets of interest, and our task is to automatically detect and localize vehicles from complex urban scenes (see Fig.~\ref{fig:FirstShow}). This is actually an exceedingly challenging task, because of 1) huge differences in visual appearance among cars (e.g., colors, sizes, and shapes) and 2) various orientations of vehicles.

\subsection{Detection of Vehicles Using Feature Engineering}
Since effective feature representation is a matter of great importance to an object detection system, traditionally, vehicle detection in remote sensing images was dominated by works that make use of low-level, hand-crafted visual features (e.g., color histogram, texture feature, scale-invariant feature transform (SIFT)~\cite{1999SIFT}, and histogram of oriented gradients (HOG)~\cite{2005HOG}) and classifiers. For example, in~\cite{Shao12}, the authors incorporate multiple visual features, local binary pattern (LBP)~\cite{2002LBP}, HOG, and opponent histogram, for vehicle detection from high resolution aerial images. Moranduzzo and Melgani~\cite{Moranduzzo141} first use SIFT to detect interest points of vehicles and then train a support vector machine (SVM) to classify these interest points into vehicle and non-vehicle categories based on the SIFT descriptors. They later present an approach~\cite{2014Moranduzzo142} that performs filtering operations in horizontal and vertical directions to extract HOG features and yield vehicle detection after the computation of a similarity measure, using a catalog of vehicles as a reference. In~\cite{2015MunichBaseline}, the authors make use of an integral channel concept, with Haar-like features and an AdaBoost classifier in a soft-cascade structure, to achieve fast and robust vehicle detection. In~\cite{2017CarDet1}, the authors use a sliding window framework consisting of four stages, namely window evaluation, extraction and encoding of features, classification, and postprocessing, to detect cars in complex urban environments by using a combined feature of the local distributions of gradients, colors, and texture. Moreover, in~\cite{2017CarDet2}, the authors utilize a multi-graph region-based matching method to detect moving vehicles in segmented UAV video frames. In~\cite{zhou2018car}, the authors apply a bag-of-words (BoW) model and a local steering kernel with the sliding window strategy to detect vehicles in arbitrary orientations, but this method is quite time consuming.

\subsection{Detection of Vehicles Using Convolutional Neural Networks}
The aforementioned methods mainly rely on manual feature engineering to build a classification system. Recently, as an important branch of deep learning family, convolutional neural networks (CNNs) have become the method of choice in many computer vision and remote sensing problems~\cite{Mou1} (e.g., object detection~\cite{han2015,han2016,2016zhangplaneDet,2017AVPN,mine2018,mine2018icassp,2018MouSeg,mine2018igarss,Xia_2018_CVPR}) as they are capable of automatically extracting mid- and high-level features from raw images for the purpose of visual analysis. For example, Chen et al.~\cite{PanLetter14} propose a vehicle detection model, called hybrid deep neural network, which consists of a sliding window technique and CNN. The main insight behind their model is to divide feature maps of the last convolutional layer into different scales, allowing for the extraction of multi-scale features for vehicle detection. In~\cite{Ammour17}, the authors segment an input image into homogeneous superpixels that can be considered as vehicle candidate regions, making use of a pre-trained deep CNN to extract features, and train a linear SVM to classify these candidate regions into vehicle and non-vehicle classes. Moreover, several recent works focus on a similar task, vehicle instance segmentation. For instance, Audebert et al.~\cite{AudebertRS17} propose a deep learning-based three-stage method called ``segment-before-detect'' for the semantic segmentation and subsequent classification of several types of vehicles in high-resolution remote sensing images. The use of SegNet~\cite{SegNet} in this method is capable of producing pixel-wise annotations for vehicle semantic mapping. Mou et al.~\cite{2018MouSeg} propose a unified multi-task learning network that can simultaneously learn two complementary tasks -- namely, segmenting vehicle regions and detecting semantic boundaries. The latter subproblem is helpful for differentiating ``touching'' vehicles, which are usually not correctly separated into instances.

\subsection{Is Non-Rotatable Detection Enough for Vehicle Detection?}
As our survey of related work shows above, most of existing approaches have focused on non-rotatable car detection~\cite{Moranduzzo141,2014Moranduzzo142,2015MunichBaseline,2016Car2,Cao2016Robust,2017CarDet1,2017Car1}, i.e., detecting all instances of vehicles and localizing them in the image in the form of horizontal bounding boxes with confidence scores. Detecting vehicles of arbitrary orientations in complex urban environments has received much less attentions and remains a challenging for most car detection algorithms, while the orientation information of vehicles is of importance for some practical applications such as traffic monitoring. In this paper, we make an effort to build an effective, rotatable detection model for vehicles of arbitrary orientations in complicated urban scenes.

\begin{figure*}[tb!]
\centering
\includegraphics[width=175mm]{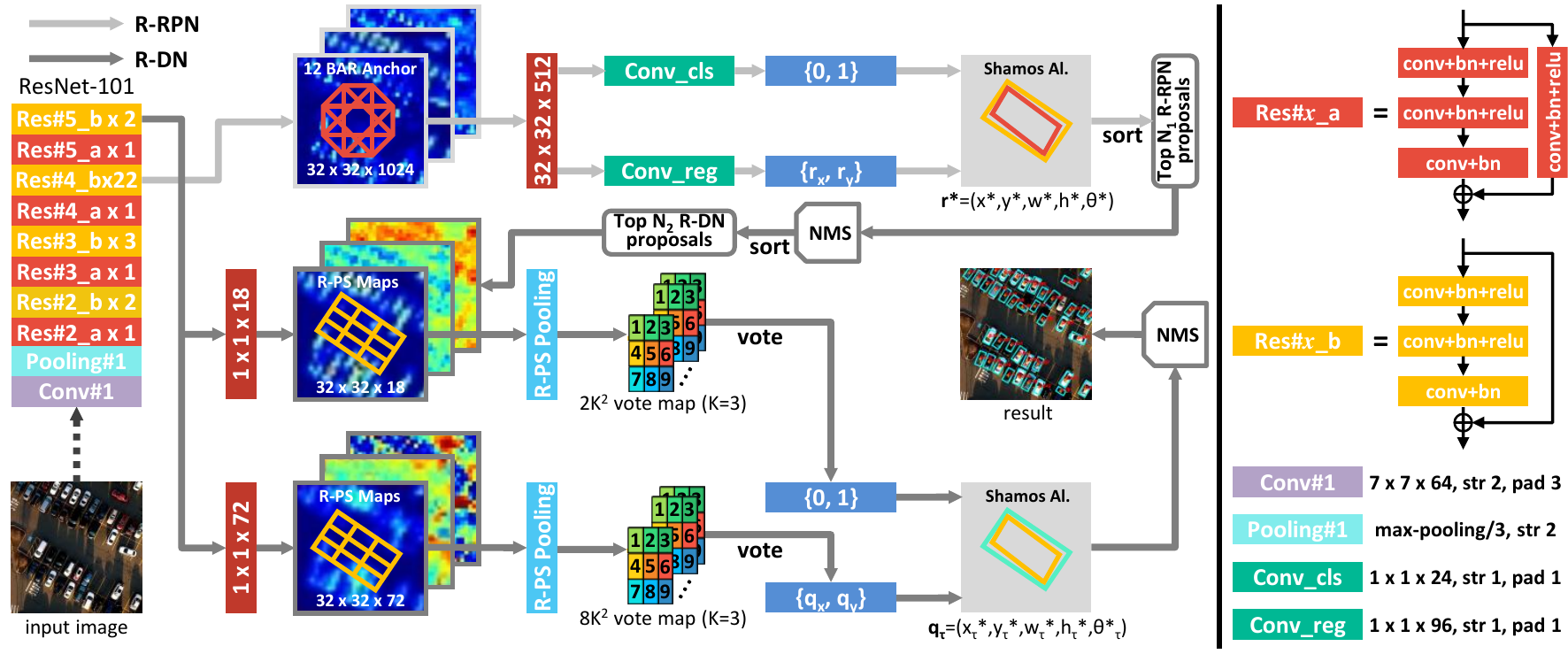}	%%%%%%%%%%LiFig02
\caption{Overall architecture of the proposed R$^3$-Net. There are two main branches. One is rotatable region proposal network (R-RPN), and the other is rotatable detection network (R-DN). R-RPN is for generating rotatable region of interests (R-RoIs) for R-DN, and then R-RoIs are fed into R-DN to produce final rotatable detection boxes. Key: str = stride; pad = padding; conv = convolution; bn = batch normalization; relu = ReLU function.}
\label{fig:R3-Net}
\end{figure*}

In this paper, we propose an end-to-end trainable network, rotatable region-based residual network (R$^3$-Net), for simultaneously localizing vehicles and identifying their orientations in high resolution remote sensing images. To this end, we introduce a series of effective rotatable operations in the network, aiming at generating multi-oriented bounding boxes for our task. When directly applied to detect vehicles of arbitrary orientations, conventional detection networks (e.g., Faster R-CNN~\cite{RenShaoqing2015FasterRCNN} and R-FCN~\cite{Dai2016RFCN}) that are primarily designed for horizontal detection would result in low precisions. In contrast, the proposed rotatable network is capable of offering better performance, especially in some complex scenes such as a crowed parking lot. In this paper, we also try to apply our network to car detection and tracking in aerial videos and find that R$^3$-Net is able to provide satisfactory vehicle trajectories. Moreover, when we take into account the temporal information of a video (i.e., the frame association produced by multiple object tracking algorithms), better detection results can be obtained. This paper contributes to the literature in the following three aspects:
\begin{itemize}
  \item We take advantage of the axial symmetry property of vehicles to create a novel network architecture, which is based on conventional two-stage object detection framework (e.g., R-FCN) but able to generate and handle rotatable bounding boxes by two tailored modules, namely rotatable region proposal network (R-RPN) and rotatable detection network (R-DN). In addition, on top of R-RPN and R-DN, we use a modified Shamos Algorithm to obtain regular quadrilaterals.
  \item We propose a novel rotatable position-sensitive RoI pooling operation, namely R-PS pooling, in order to reduce the dimension of feature maps of rotatable regions and meanwhile, keep the information of targets in specific directions.
  \item We propose a novel strategy, called BAR anchor, to initialize rotatable anchors of region proposals using the size information of vehicles in the training set. Experimental results show that this strategy is able to estimate vehicle poses more accurately as compared to traditional anchor generation method.
\end{itemize}

The remainder of this paper is organized as follows. After the introductory Section~\ref{sec:introduction}, detailing vehicle detection from high resolution remote sensing imagery, we enter Section~\ref{sec:method}, dedicated to the details of the proposed R$^3$-Net for multi-oriented vehicle detection. Section~\ref{sec:experiments} then provides dataset information, implementation settings, and experimental results. Finally, Section~\ref{sec:conclusion} concludes the paper.

\section{Methodology}% 3.
\begin{figure}[tb!]
\centering
\includegraphics[width=88mm]{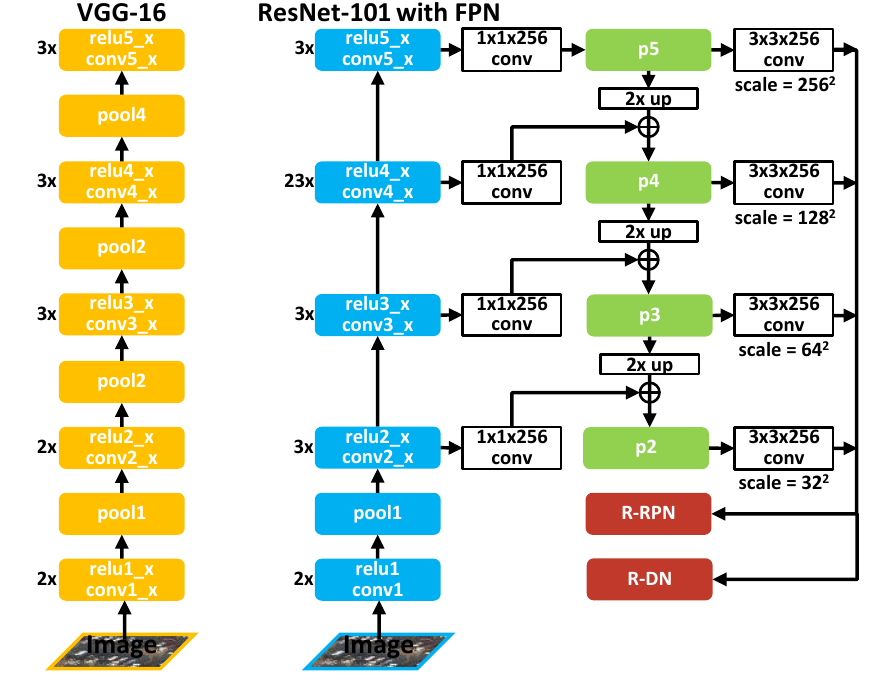}
\caption{Two typical network architectures: VGG-16 and ResNet-101 with feature pyramid network (FPN).}
\label{fig:FEN}
\end{figure}

\label{sec:method}
Our approach for vehicle detection utilizes an end-to-end trainable two-stage detection framework, including a rotatable region proposal network (R-RPN) and a rotatable detection network (R-DN). As shown in Fig.~\ref{fig:R3-Net}, after the feature extraction by a ResNet-101~\cite{Hekaiming2016ResNet}, we devise the framework in a rotatable detection domain to preserve competitive detection rates, and the details are elaborated in the following subsections.
Besides, we also utilize two other typical deep networks for feature extraction, VGG-16~\cite{Simonyan2014VGG} and ResNet-101 with feature pyramid network (FPN)~\cite{FPN2016ROSS} (see Fig.~\ref{fig:FEN}).

\subsection{Rotatable region proposal network}% 3.3
The main task of R-RPN is to generate cursory rotatable region of interests (R-RoIs) for the subsequent thorough detection. Referring to the strategy of generating region of interests (RoIs) in Faster R-CNN~\cite{RenShaoqing2015FasterRCNN}, we propose to take batch averaging rotatable anchors (BAR anchors) as the input of R-RPN and produce a series of R-RoIs as output. We next describe the main ingredients of R-RPN.

\textbf{BAR Anchors.} In the RPN stage of two-stage detection methods, such as Faster R-CNN~\cite{RenShaoqing2015FasterRCNN} and R-FCN~\cite{Dai2016RFCN}, anchors are initial shapes of RoIs on each feature point, and the pattern of the anchors contributes to the pattern of RoIs. For instance, an anchor in the shape of rectangle finally regresses to an RoI of rectangle shape. Likewise, an R-RoI can be obtained by a rotatable anchor, and the size of anchors depends on the size of detected objects.

With regard to the vehicle detection task from remote sensing images, we use anchors of permanent size in each training mini-batch for the mostly invariant overall dimension of vehicle. Formally, we consider that there are $M$ mini-batches in the training process, and batch size is set to $N_m$. We use $\mathcal{I}^{(m)}_{i}$ to represent the $i$-th ($i = 1,...,N_m$) training sample in the $m$-th ($m = 1,...,M$) mini-batch, so the $j$-th rotatable ground truth box in $\mathcal{I}^{(m)}_{i}$ can be represented by a coordinate $(x_{ij}, y_{ij}, w_{ij}, h_{ij}, \theta_{ij})_{m}$. Then we define the average width $\hat{w}_m$ and height $\hat{h}_m$ of initialized anchors in $m$-th training mini-batch as follows:
\begin{equation}
\begin{split}
(\hat{w}_m,\hat{h}_m) &= \frac{1}{\sum_{i=1}^{N_m} N_i} \sum_{i=1}^{N_m} \sum_{j=1}^{N_i} (w_{ij},h_{ij})\,,   \label{Eq:1}
\end{split}
\end{equation}
where $N_i$ is the number of rotatable ground truth boxes in $\mathcal{I}^{(m)}_{i}$. Hence, for the $m$-th training mini-batch, the coordinate of BAR anchor on feature point $(x,y)$ can be defined as $\mathbf{a^{*}} = (x,y,\kappa \hat{w}_m,\kappa \hat{h}_m,\theta)$, where $\theta \in \{ -45^\circ, 0^\circ, 45^\circ, 90^\circ \}$ and scale factor $\kappa$ is set to $\{0.5,1,2\}$ in this paper. As a result, we get 12 BAR anchors on each feature point.

\begin{figure*}[tb!]
\centering
\includegraphics[width=175mm]{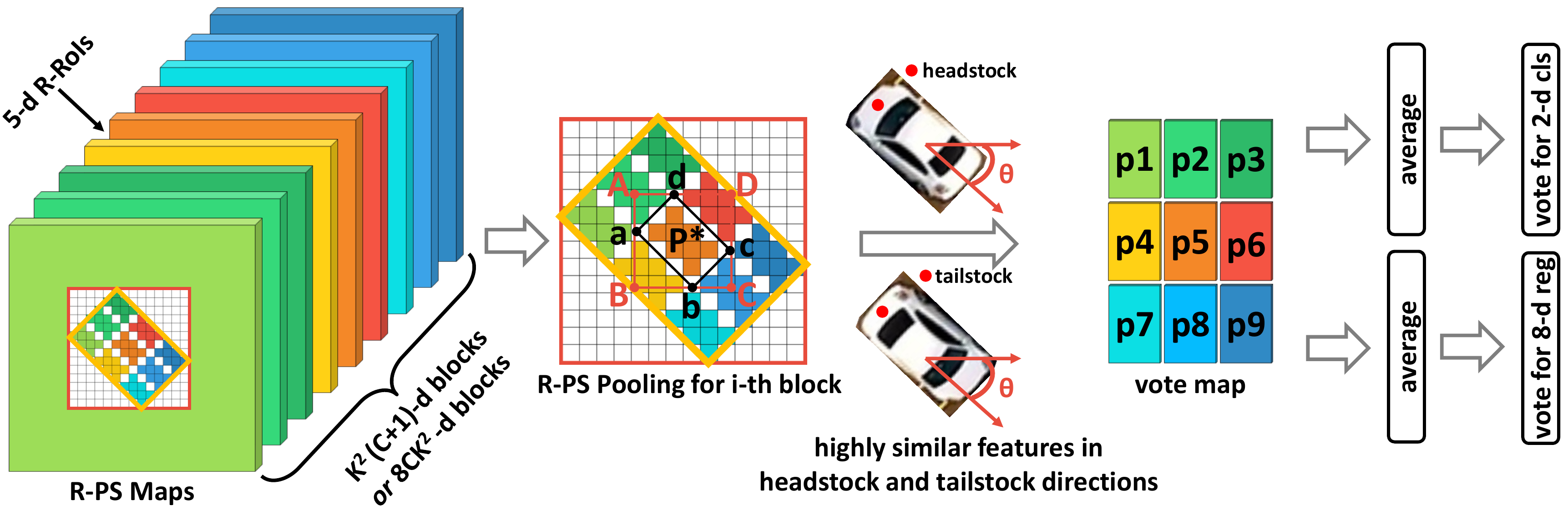}	%%%%%%%%%%LiFig02
\caption{Details of rotatable position sensitive pooling (R-PS pooling) operation.}
\label{fig:4}
\end{figure*}

\textbf{Definition of Parameters in Regression.} Unlike traditional object detection networks that generate horizontal RoIs and represent them by 4-d vectors (i.e., an RoI's center coordinates and its width and height), the proposed method needs to produce R-RoIs, which should be defined by 5-d vectors, namely an RoI's center coordinates $(x,y)$, width $w$, height $h$, and intersection angle $\theta$ ($-90^{\circ}< \theta \leq 90^{\circ}$), between its lengthwise direction and horizontal direction. However, in our experiments, we found that the use of a 8-d vector for representing R-RoI box $\mathbf{r} = (\mathbf{r_x},\mathbf{r_y})$ and ground truth box $\mathbf{g} = (\mathbf{g_x},\mathbf{g_y})$, where $\mathbf{r_x} = (x_{1},...,x_{4})$, $\mathbf{r_y} = (y_{1},...,y_{4})$, $\mathbf{g_x} = (x_{g1},...,x_{g4})$, and $\mathbf{g_y} = (y_{g1},...,y_{g4})$, can make regression loss easier to be optimized. This is mainly because the 8-d vector-based way is capable of alleviating the unsteadiness caused by parameter $\theta$ when computing the regress loss. The following experimental results will show different influences of these two definitions.

To match with the dimension of R-RoI, we also convert the BAR anchors into the 8-d vectors from $\mathbf{a}$ as follows:
\begin{equation}
\begin{split}
\mathbf{a} = G(\mathbf{a^{*}}) = (\mathbf{a_x},\mathbf{a_y})\,,  \label{Eq:2}
\end{split}
\end{equation}
where $G(\cdot)$ is a biunique geometric transformation, $\mathbf{a_x} = (x_{a1},...,x_{a4})$, and $\mathbf{a_y} = (y_{a1},...,y_{a4})$. Notably, the collation of these 4 vertexes in each BAR anchor's representation $\mathbf{a}$ should be rigorously matched with those in ground truth labels and R-RoIs, and the collation rule of all rotatable rectangles is defined as follows:
\begin{enumerate}
  \item Confirm the intersection angle $\alpha$ between its lengthwise direction and horizontal direction ($-90^{\circ}< \alpha \leq 90^{\circ}$);
  \item Rotate the rectangle around its center to the horizontal direction by $-\alpha$;
  \item Label four vertexes in the order of coordinate.
\end{enumerate}

Consequently, in bounding box regression step, for the $n$-th feature point $f_n$ ($n= 1,...,N$, e.g., $N = 32\times32$ in ResNet-101), we define ${\mathbf{t}}^{(k)}_{n}=(\mathbf{t_x},\mathbf{t_y})^{(k)}_{n}$ as an 8-d vector representing eight parameterized coordinates of a predicted bounding box, and $\mathbf{\hat{t}}_{n}^{(k)}=(\mathbf{\hat{t}_x},\mathbf{\hat{t}_y})_{n}^{(k)}$ is its corresponding ground truth box where $k$ ($k$ = 1,...,12) is the index of a BAR anchor on the feature point $f_n$. Then we define parameterizations of four coordinates as follows~\cite{Ross2013RCNN}:
\begin{equation}
\begin{split}
\mathbf{t_x}&=\frac{\mathbf{r_x}-\mathbf{a_x}}{\kappa \hat{w}_m}, \quad \mathbf{t_y}=\frac{\mathbf{r_y}-\mathbf{a_y}}{\kappa \hat{h}_m}\,,\\
\mathbf{\hat{t}_x}&=\frac{\mathbf{g_x}-\mathbf{a_x}}{\kappa \hat{w}_m}, \quad \mathbf{\hat{t}_y}=\frac{\mathbf{g_y}-\mathbf{a_y}}{\kappa \hat{h}_m}\,.  \label{Eq:3}
\end{split}
\end{equation}

\textbf{Multi-Task Loss.} In common with other object detection networks, such as Faster R-CNN~\cite{RenShaoqing2015FasterRCNN} and R-FCN~\cite{Dai2016RFCN}, we define a multi-task loss to combine both classification and regression loss together. The loss function in R-RPN is defined as an effective multi-task loss $L_{1}$~\cite{Ross2015FastRCNN}, which combines both classification loss and regression loss for each image. It can be computed as follows
\begin{equation}
\begin{split}
L_{1} & =  \frac{1}{N_{cls}} \sum_{n,k} L_{cls}({\mathbf{p}_{n}^{(k)}},{\mathbf{\hat{p}}_{n}^{(k)}}) \\
                     & + \lambda_1 \frac{1} {N_{reg}} \sum_{n,k} \phi_{n}^{(k)} L_{reg}(\mathbf{t}_{n}^{(k)},\mathbf{\hat{t}}_{n}^{(k)})\,, \label{Eq:4}
\end{split}
\end{equation}
where ${\mathbf{p}}_{n}^{(k)}$ $= (p^0,p^1)_n^{(k)}$ is the predicted probability of the $k$-th ($i=1,...,12$) anchor in the $n$-th ($n=1,...,1024$) feature point being background and a target, and ground truth indicator label $\mathbf{\hat{p}}_{n}^{(k)}$ is $(0,1)$ if the anchor is positive (i.e., \emph{overlap} $\geq$ 0.5), and is $(1,0)$ if the anchor is negative. Ground truth indicator label $\phi_{n}^{(k)}$ is 1 when there exist targets in the anchor.

We define the classification loss $L_{cls}$ as $\log$ loss over two classes (background and target) as follows:
\begin{equation}
\begin{split}
L_{cls}(\mathbf{p}_{n}^{(k)},\mathbf{\hat{p}}_{n}^{(k)}) = -\mathbf{p}_{n}^{(k)} \log \mathbf{p}_{n}^{(k)}\,. \label{Eq:5}
\end{split}
\end{equation}

Then we define the regression loss $ L_{reg} $ as
\begin{equation}
\begin{split}
L_{reg}(\mathbf{t}_{n}^{(k)},\mathbf{\hat{t}}_{n}^{(k)}) = R(\mathbf{t}_{n}^{(k)}-\mathbf{\hat{t}}_{n}^{(k)})\,, \label{Eq:6}
\end{split}
\end{equation}
where $R(\cdot)$ is a smooth $\ell_1$ loss function which is defined in~\cite{Ross2015FastRCNN}, and formally, it can be calculated by

\begin{equation} \label{Eq:7}
R(x)=
\begin{cases}
0.5x^{2} &\mbox{if $|x|<1$}\\
|x|-0.5 &\mbox{otherwise}
\end{cases}\,.
\end{equation}

In Eq.~\myref{Eq:4}, two loss terms are normalized by $N_{cls}$ and $N_{reg}$ and balanced by hyper-parameter $\lambda_1$. In our experiments, we set $N_{cls}=64$, $N_{reg}=1000$, and $\lambda_1=10$.

\textbf{Shamos Algorithm for R-RoIs.} The rotating calipers method was first used in the dissertation of Michael Shamos in 1978~\cite{shamos1978computational}. Shamos used this method to generate all antipodal pairs of points on a convex polygon and to compute the diameter of a convex polygon in $\mathcal{O}(n)$ time. Then Houle and Toussaint developed an application for computing the minimum width of a convex polygon in~\cite{1988Shamos}. After R-RPN, we actually obtain R-RoIs in the form of irregular quadrilateral denoted by 8-d vectors. In order to get R-RoIs in the form of regular quadrilateral to feed them into R-DN, here, we propose to use Shamos Algorithm to calculate the minimum multi-oriented rectangular bounding boxes. The coordinate transformation can be described as follows:
\begin{equation}
\begin{split}
\mathbf{r^{*}} = (x^*,y^*,w^*,h^*,\theta ^*) = S(\mathbf{r}) \,,  \label{Eq:8}
\end{split}
\end{equation}
where a 5-d vector $(x^*,y^*,w^*,h^*,\theta ^*)$ is utilized to represent a minimum multi-oriented rectangular R-RoI box $\mathbf{r^{*}}$, which can be obtained from an 8-d vector of an irregular quadrilateral R-RoI box $\mathbf{r}$ by the Shamos Algorithm $S(\cdot)$. Now we can feed the rectangular R-RoI boxes into R-DN.

\subsection{Rotatable detection network}
Suppose that we generate $T$ R-RoIs in total after R-RPN, and these R-RoIs are subsequently fed into R-DN for the final regression and classification tasks. Here, an improved position-sensitive RoI pooling strategy, called rotatable position-sensitive RoI pooling (R-PS pooling), is proposed to generate scores on rotatable position-sensitive score maps (R-PS maps) for each R-RoI. We next describe the main ingredients of our R-DN.

\textbf{R-PS Maps of R-RoIs.} Given R-RoIs defined by 5-d vectors, each target proposal can be located on the feature maps extracted from an adjusted ResNet-101~\cite{Hekaiming2016ResNet}, which uses a randomly initialized $1\times 1$ convolutional layer with 1024 filters instead of a global average pooling layer and a fully connected layer. The size of output feature maps is $32 \times 32 \times 1024$.

For the classification task in R-DN, we apply $k^2$ R-PS maps for each category and $k^2(C+1)$-channel output layer with $C$ object categories ($C=1$ for our vehicle detection task and $+1$ for background). The bank of $k^2$ R-PS maps corresponds to a $k \times k$ spatial grid describing relative positions, and we set $k=3$ in this paper.

Different from position-sensitive score maps in R-FCN~\cite{Dai2016RFCN}, the R-PS maps in our method do not encode cases of an optional object category, but cases of a potential vehicle category. As shown in Fig.~\ref{fig:4}, R-RoIs of vehicles are always approximate central symmetric so that there are always highly similar features in headstock and tailstock direction for most vehicles, making it hard to identify the exact direction. In order to avoid this almost "unavoidable" misidentification, in our model, the angle $\theta^*$ of an R-RoI $\mathbf{r^*}$ is kept spanning $-90^{\circ}< \theta^* \leq 90^{\circ}$. For example, $\theta^* = -45^\circ$ shows two possible cases that the vehicle direction might be $-45^\circ$ or $135^\circ$, and these two possible cases are encoded into R-PS maps in the same order.

\textbf{R-PS Pooling on R-PS Maps.} We divide each R-RoI rotatable rectangular box into $3 \times 3$ bins by a parallel grid. For an R-RoI $\mathbf{r^*} = (x^*,y^*,w^*,h^*,\theta ^*)$ ($w^*\leq h^*$), a bin is of size $\approx w^*/3 \times h^*/3$~\cite{Ross2015FastRCNN, HeKaiming2015SPPNet}. For the $(i,j)$-th bin ($i,j=1,2,3$), the R-PS pooling step over the $(i,j)$-th R-PS map of $c$-th category ($c=1,2$) is defined as follows:
\begin{equation}
\begin{split}
r_{i,j,c}(\mathbf{w})= \sum_{(u,v)\in \mathbf{B}_{i,j}} \frac{1}{n_p}z_{i,j,c}(u,v|\mathbf{w})\,, \label{Eq:9}
\end{split}
\end{equation}
where $r_{i,j,c}$ denotes pooled output in the $(i,j)$-th bin $\mathbf{B}_{i,j}$ for the $c$-th category, $z_{i,j,c}$ represents one R-PS map out of the $k^2(C+1)$ score maps, $\mathbf{w}$ is all learnable parameters of the network, $n_p$ is the number of pixels in the $(i,j)$-th bin $\mathbf{B}_{i,j}$, and $(u,v)$ is the global coordinate of feature point $P_{i,j}$ which can be defined by the following affine transformation equation:
\begin{equation}
\begin{bmatrix}
        u \\
        v \\
        1 \\
\end{bmatrix}=
\begin{bmatrix}
  \cos \hat{\theta} & -\sin \hat{\theta} & u_0 \\
  \sin \hat{\theta} & \cos \hat{\theta} & v_0 \\
  0 & 0 & 1 \\
\end{bmatrix}
\begin{bmatrix}
  \Delta u\\
  \Delta v \\
  1 \\
\end{bmatrix}\,, \label{Eq:10}
\end{equation}
where $(\Delta u, \Delta v)$ means local coordinates of feature point $P_{i,j}$ and $ \lfloor (i-1) w^*/3 \rfloor \leq \Delta u < \lceil i w^*/3 \rceil, \lfloor (j-1) h^*/3 \rfloor \leq \Delta v < \lceil j h^*/3 \rceil$, and $(u_0, v_0)$ means the top-left corner of an R-RoI. Formally, the rotation angle $\hat{\theta}$ can be caculated by

\begin{equation}
\hat{\theta}=
\begin{cases}
\theta^*-90^{\circ} &\mbox{$-90^{\circ} < \theta^* \leq 0^{\circ}$}\\
90^{\circ}-\theta^* &\mbox{$0^{\circ} < \theta^* \leq 90^{\circ}$}
\end{cases}\,. \label{Eq:11}
\end{equation}

When the $(i,j)$-th bin $\mathbf{B}_{i,j}$ is pooled into a score map, we can get $(\Delta u, \Delta v)$ by $(u, v)$ as follows:
\begin{equation}
\begin{split}
\Delta u &= u\cos \hat{\theta}-v \sin \hat{\theta}-u_0 \,,\\
\Delta v &= v \cos \hat{\theta} -u\sin \hat{\theta} -v_0 \,,\label{Eq:12}
\end{split}
\end{equation}
and we then get the limitation of $u$ and $v$ when $(u,v) \in \mathbf{B}_{i,j}$ in the following inequalities:
\begin{equation}
\begin{split}
 \lfloor (i-1) w^*/3 &\rfloor \leq u\cos \hat{\theta}-v \sin \hat{\theta}-u_0 < \lceil i w^*/3 \rceil \,, \\
 \lfloor (j-1) h^*/3 &\rfloor \leq v \cos \hat{\theta}-u\sin \hat{\theta} -v_0 < \lceil j h^*/3 \rceil \,. \label{Eq:13}
\end{split}
\end{equation}

\textbf{Voting by Scores.} After performing R-PS pooling operation on R-PS maps in $3\times3$ positions with 2 layers, we can then puzzle the $3\times3$ blocks into one voting map for each R-RoI. Here, a voting map is a kind of feature map that can keep rotatable position-sensitive features on different position blocks. Hence, we use total outputs $r_c(\mathbf{w})$ on $3\times3$ blocks to compute the score of the $c$-th category by
\begin{equation}
\begin{split}
r_c(\mathbf{w}) = \sum_{i,j} r_{i,j,c}(\mathbf{w}), \quad i,j=1,2,3\,. \label{Eq:14}
\end{split}
\end{equation}

And the softmax response of the $c$-th category ($c=1,2$) can be computed as follows:
\begin{equation}
\begin{split}
s_c(\mathbf{w}) = \frac {e^{r_c(\mathbf{w})}} {\sum_{\delta} e^{r_\delta(\mathbf{w})}},  \quad \delta=1,2\,. \label{Eq:15}
\end{split}
\end{equation}

\textbf{Loss Function in R-DN.} Similarly, for regression task in R-DN, we use an 8-d vector to represent eight parameterized coordinates of a predicted bounding box and apply $3\times3$ R-PS maps to each dimension for regression. Thus, the R-PS score maps are fed into a $1 \times 1$ convolutional layer with 72 filters for bounding box regression. Then we pool these feature maps into a 72-d vector which is aggregated into an 8-d vector by average voting for the $\tau$-th ($\tau = 1,...,T$) predicted bounding box ${\mathbf{q}_{\tau}}=(\mathbf{q_x},\mathbf{q_y})_{\tau}$.

Likewise, we also define a multi-task loss $L_{2}$ for $N$ RoIs:
\begin{equation}
\begin{split}
L_{2} & = \sum_\tau L(\mathbf{s}_\tau, \mathbf{\hat{p}}_\tau, \mathbf{q}_\tau, \mathbf{\hat{t}}_\tau) \\
       & = \sum_\tau L_{cls}(\mathbf{s}_\tau,\mathbf{\hat{p}}_\tau) + \lambda_2 \sum_\tau \phi_\tau L_{reg}(\mathbf{q}_\tau,\mathbf{\hat{t}}_\tau)\,, \label{Eq:16}
\end{split}
\end{equation}
where $\mathbf{\hat{p}}_\tau$ represents ground truth (e.g., 0 and 1 stand for background and vehicle, respectively), and ground truth indicator label $\phi_\tau$ is 1 when there exist vehicles in the $\tau$-th ($\tau=1,...,T$) R-RoI's predicted box (i.e., overlap $\geq$ 0.5). $\mathbf{s}_\tau$ is the predicted score of the R-RoI being a vehicle, and $\mathbf{s}_\tau=(s_\tau^{1},s_\tau^{2}) $ covers two categories. As usual, we use a fully connected layer activated by a softmax function to compute $\mathbf{s}_\tau$ in Eq.~\myref{Eq:15}. Moreover, ${\mathbf{q}_{\tau}}=(\mathbf{q_x},\mathbf{q_y})_\tau $ and $\mathbf{\hat{t}}_\tau=(\mathbf{\hat{t}_x},\mathbf{\hat{t}_y})_\tau$ represent coordinates of the predicted bounding box and ground truth box, which are similarly parameterized by Eq.~\myref{Eq:3}. In addition, we define the classification loss $L_{cls}$ as $\log$ loss for two classes like Eq.~\myref{Eq:5} and the regression loss $L_{reg}$ as smooth $\ell_1$ loss referring to Eq.~\myref{Eq:6} and Eq.~\myref{Eq:7}. In Eq.~\myref{Eq:16}, two terms of the loss are also normalized by a hyper-parameter, i.e., $\lambda_2$. By default, we set $\lambda_2=1$.

\subsection{Joint loss function}
The joint loss function $ L $ of our end-to-end trainable two-stage detection framework is a combination of R-RPN loss $L_{1}$ and R-DN loss $L_{2}$, and we use a loss weight $\eta$ to balance them. For each mini-batch (batch size $\Theta=64$ in our experiment), we compute the joint loss as follows:
\begin{equation}
\centering
\begin{split}
L = \sum_{\theta}^{\Theta}L_{1}^{(\theta)}+\eta\sum_{\theta}^{\Theta}L_{2}^{(\theta)}+\varphi\parallel\mathbf{w}\parallel^2 \,, \label{Eq:17}
\end{split}
\end{equation}
where $L_{1}^{(\theta)}$, and $L_{2}^{(\theta)}$ can be calculated by Eq.~\myref{Eq:2} and Eq.~\myref{Eq:6} for the $\theta$-th ($\theta=1,2,...,\Theta$) image in a mini-batch. In this paper, we set the weight $\eta=1$. We also conducted experiments to find better $\eta$ out, and the details will be given in~Section~\ref{sec:experiments}.

\begin{figure}[t!]
\centering
\includegraphics[width=88.8mm]{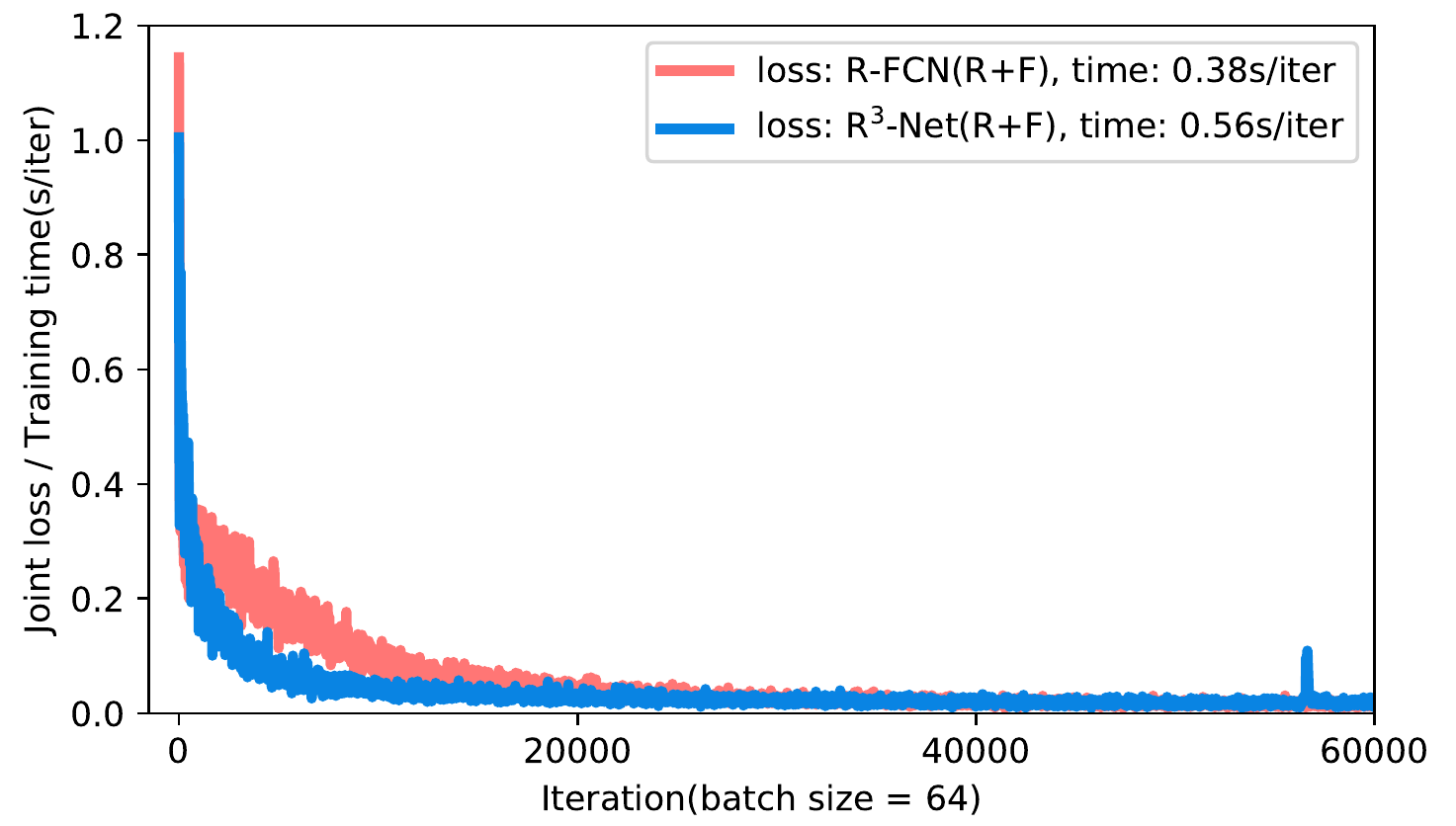}
\caption{The joint loss and training time curves of R-FCN and R$^3$-Net. Key: (R+F) = ResNet-101 with FPN.}
\label{fig:JointLoss}
\end{figure}

\subsection{End-to-end training}
An end-to-end training strategy is utilized to train our model, and we use a mini-batch gradient descent algorithm to update network weights $\mathbf{w}$. Here, we define $\mathbf{w}=\{ \mathbf{w}_{res}, \mathbf{w}_{rrpn},\mathbf{w}_{rdn}\}$ to represent learnable parameters of ResNet-101, R-RPN, and R-DN, respectively, and $\mathbf{w}_{res}=\{ \mathbf{w}_{res\_1},...,\mathbf{w}_{res\_5}\}$ indicates parameters of five blocks in ResNet-101. In this paper, to train our model more efficiently, we only update parameters of the last two blocks of ResNet-101 and keep those of the first three blocks fixed. There are two training stages for R-RPN and R-DN, respectively. In R-RPN, feature maps are extracted by the first four blocks of ResNet-101, and $L_{1}^{(\theta)}$ is computed and accumulated in a mini-batch $\Theta$. In R-DN, we extract feature maps using all five blocks, compute $L_{2}^{(\theta)}$ of an image, and accumulate it in a mini-batch $\Theta$. We then compute the joint loss $L$ by Eq.~\myref{Eq:17} and independently perform back-propagation at the end of each mini-batch.

As aforementioned, there are two transmission routes which are regarded as forward propagation in our network. We initialize the network parameters $\{\mathbf{w}_{res\_4},\mathbf{w}_{res\_5}\}$ and $\{\mathbf{w}_{rrpn},\mathbf{w}_{rdn}\}$ by pre-trained model and Gaussian distribution, respectively. At the end of R-RPN and R-DN, according to the discrepancy between ground truth and the stepwise output of R-RPN and R-DN, we can use gradient descent algorithm to update learnable parameters of different parts with a learning rate $\varepsilon$ as follows:
\begin{equation}
\begin{split}
&\mathbf{w}_{rrpn} := \mathbf{w}_{rrpn} - \varepsilon \sum^{\Theta}_{\theta}\frac{dL_1^{(\theta)}}{d\mathbf{w}_{rrpn}}\ + 2\varphi\parallel\mathbf{w}_{rrpn}\parallel\,, \\ \label{Eq:18}
&\mathbf{w}_{rdn} := \mathbf{w}_{rdn} - \varepsilon \sum^{\Theta}_{\theta}\frac{\eta dL_2^{(\theta)}}{d\mathbf{w}_{rdn}}\ + 2\varphi\parallel\mathbf{w}_{rdn}\parallel\,, \\
&\mathbf{w}_{res\_5} := \mathbf{w}_{res\_5} - \varepsilon \sum^{\Theta}_{\theta}\frac{dL_1^{(\theta)}}{d\mathbf{w}_{res\_5}}\ + 2\varphi\parallel\mathbf{w}_{res\_5}\parallel\,, \\
&\mathbf{w}_{res\_4} := \mathbf{w}_{res\_4} - \varepsilon  \sum^{\Theta}_{\theta}\frac{dL_1^{(\theta)}+\eta dL_2^{(\theta)}}{d\mathbf{w}_{res\_4}}\ + 2\varphi\parallel\mathbf{w}_{res\_4}\parallel\,.
\end{split}
\end{equation}

\textbf{Hyper-parameter settings.} Prior to network training, all new layers are randomly initialized by drawing weights from a zero-mean Gaussian distribution with standard deviation 0.01, and all other layers are initialized by a pre-trained model on ImageNet~\cite{Russakovsky2015PASCALVOC}. In the mini-batch gradient descent, we use a learning rate $\varepsilon$ of 0.001 for the first 10K iterations and 0.0001 for the next 10K iterations on the dataset. The momentum and weight decay are set to 0.9 and 0.0005, respectively. As a result, we find that it works well after about 10K iterations. The mini-batch size $\Theta$ is set to 32 in this paper.

\section{Experiments}
\label{sec:experiments}
We use VGG-16~\cite{Simonyan2014VGG}, ResNet-101~\cite{Hekaiming2016ResNet} and ResNet-101 with FPN~\cite{FPN2016ROSS} to extract features. The models are implemented using Caffe and Caffe 2~\cite{Jia2014Caffe} and run on an NVIDIA GeForce GTX1080Ti with 12 GB on board memory.

\subsection{Dataset}
To evaluate the performance of our method, we use two open vehicle detection datasets, namely DLR 3K Munich Vehicle Dataset~\cite{2015MunichBaseline} and VEDAI Vehicle Dataset~\cite{2015VEDAIBaseline}, in which vehicles are accurately labelled by rotatable rectangular boxes. In our experiments, we regard various types of vehicles as one category. A statistic of the two datasets for our experiments can be found in Table~\ref{table:dataset}. In addition, we make use of data augmentation (translation transform, scale transform, and rotation transform) to extend the number of training samples (cf. Table~\ref{table:dataset}).

\begin{figure*}[t]
\centering
\subfigure{
\includegraphics[width=88mm]{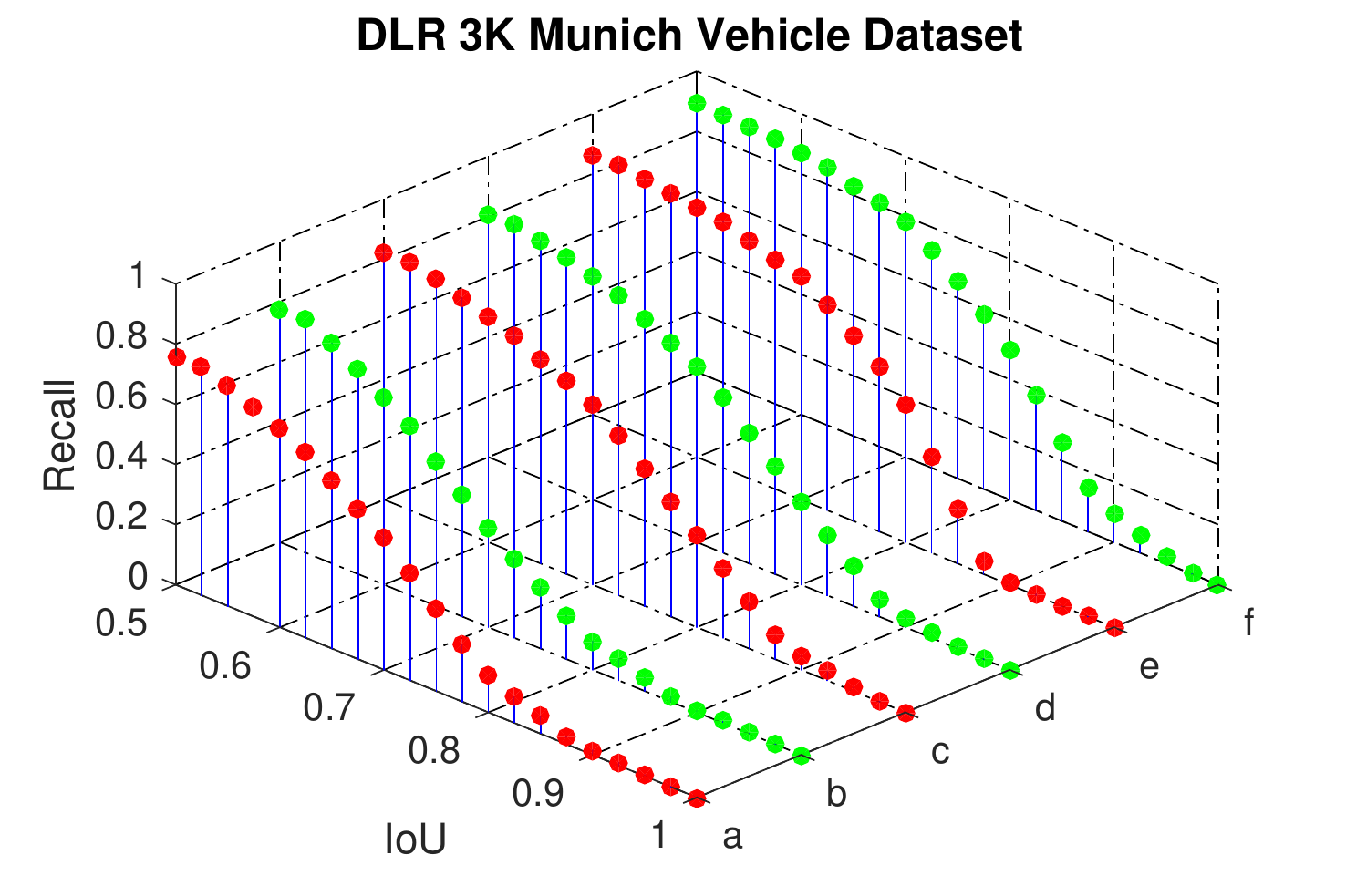}}%c--->ship
\subfigure{
\includegraphics[width=88mm]{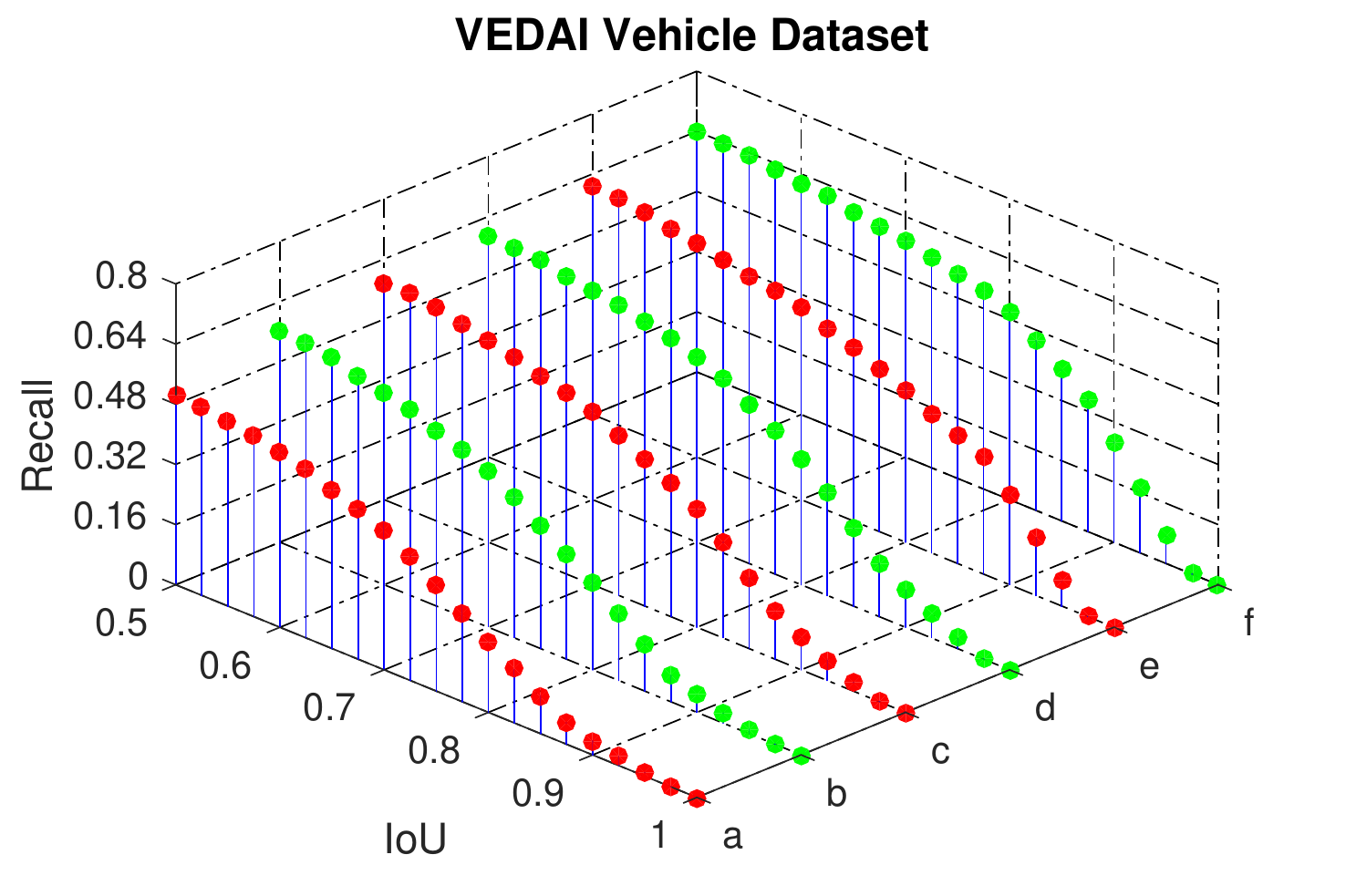}}%b--->static ship
\vspace{-2mm}
\caption{Recall-IoU comparisons between the baseline method R-FCN and the proposed R$^3$-Net. Key: a = R-FCN with VGG-16; b = R$^3$-Net with VGG-16; c = R-FCN with ResNet-101; d = R$^3$-Net with ResNet-101; e = R-FCN with ResNet-101 and FPN; f = R$^3$-Net with ResNet-101 and FPN.}
\label{fig:recall} %% label for entire figure
\end{figure*}

\subsection{Experimental analysis}
%%%%%%%%%%%%%%%%%%%%%%%%%%%%%%%%%%%%%%%%%%%%%%%%%%%%%%%%
In our detection task, there are two outputs, rotatable rectangular bounding boxes and categories (i.e., vehicle or not). In general, we evaluate the performance of detection methods by using different choices of intersection over union (IoU) which indicates the overlap ratio between a predicted box and its ground truth box. The IoU can be defined as:
\begin{equation}
\begin{split}
\mbox{IoU} = (S_{bbox}\bigcap S_{gt})/(S_{bbox}\bigcup S_{gt})\,,  \label{Eq:19}
\end{split}
\end{equation}
where $S_{bbox}$ and $S_{gt}$ are areas of the predicted box and the ground truth box, respectively, in the shape of regular rectangle. Therefore, we convert our predicted rotatable rectangular bounding boxes and rotatable rectangular ground truth boxes into regular rectangular ones (i.e., minimum bounding rectangles) for the purpose of calculating IoUs.

Then, average precision (AP) and precision recall curve are applied to evaluate object detection methods. Quantitatively, AP means the average value of precision for each object category from recall = 0 to recall = 1. For computing AP value, we define and count true positives (TPs), false positives (FPs), false negatives (FNs), and true negatives (TNs) in detection results.

For the vehicle detection task, we can regard the region of a regular rectangle bounding box as a TP in the case that the IoU is more than the given threshold value. Otherwise, if the IoU is less than the given threshold, the region is considered as an FP (also called false alarm). Moreover, the region of a target is regarded as an FN (also called miss alarm) if no predicted bounding box covers it. Otherwise, we regard the region as a TN (also called correct rejection). Consequently, we use the following definitional equations to formulate precision and recall indicators:
\begin{equation}
\begin{split}
\mbox{Precision}&=\frac{\mbox{TP}}{\mbox{TP}+\mbox{FP}} \\
\mbox{Recall}&=\frac{\mbox{TP}}{\mbox{TP}+\mbox{FN}}\,. \label{Eq:20}
\end{split}
\end{equation}

\begin{table}[t]
\centering
\caption{Dataset Overview}
\begin{tabular}{l cc }
\toprule[1pt]
\toprule[1pt]
%\cline{2-9}
    Item          & DLR 3K Munich Dataset  & VEDAI Dataset \\  %yi lie quan bu \centering cai xing
\toprule[1pt]
 \multirow{2}{*}{Category} & Car, Bus  &  Car, Pickup \\
                                & Truck     &  Truck, Van\\
 \toprule[1pt]
 Image size                    &  $512\times512$                            & $512\times512$ \\
 Data tpye                     &   RGB                                      & RGB \& NIR\\
 \toprule[1pt]
 Total image (Tr. / Te.)                     & 575 / 408 \centering               & 1066 / 1066 \\
 Original veh.(Tr. / Te.)              & 5214 / 3054  \centering            & 2792 / 2702 \\
 Augment veh.(Tr. / Te.)              &31284 / 3054  \centering             & 16752 / 2702 \\
\toprule[1pt]
\toprule[1pt]
\end{tabular}
\label{table:dataset}
\end{table}

Besides, we use F$_1$-score to evaluate the comprehensive performance of precision and recall, which can be calculated as follows:
\begin{equation}
\begin{split}
\mbox{F$_1$}&=\frac{2\times\mbox{Precision}\times\mbox{Recall}}{\mbox{Precision}+\mbox{Recall}}\,. \label{Eq:21}
\end{split}
\end{equation}

\textbf{Recall-IoU Analysis.} Fig.~\ref{fig:recall} shows Recall-IoU comparisons of our method and other baseline methods. Though a lower IoU usually means more TPs and less FPs, it may give inaccurate locations of targets at the same time. So, to obtain a better trade-off between detection and location accuracies, we habitually set the IoU threshold to 0.5. Fig.~\ref{fig:recall} also displays Recall-IoU trends of three different feature extraction networks (i.e., VGG-16, ResNet-101, and ResNet-101 with FPN). We can see that the proposed method significantly obtains better performance as compared to other networks, which indicates that the proposed network is capable of learning more robust feature representations for multi-oriented vehicle detection tasks. Besides, it can be seen that FPN can generate better localization results as it can embed low-level features into high-level ones.

\begin{figure}[tb!]
\centering
\includegraphics[width=88mm]{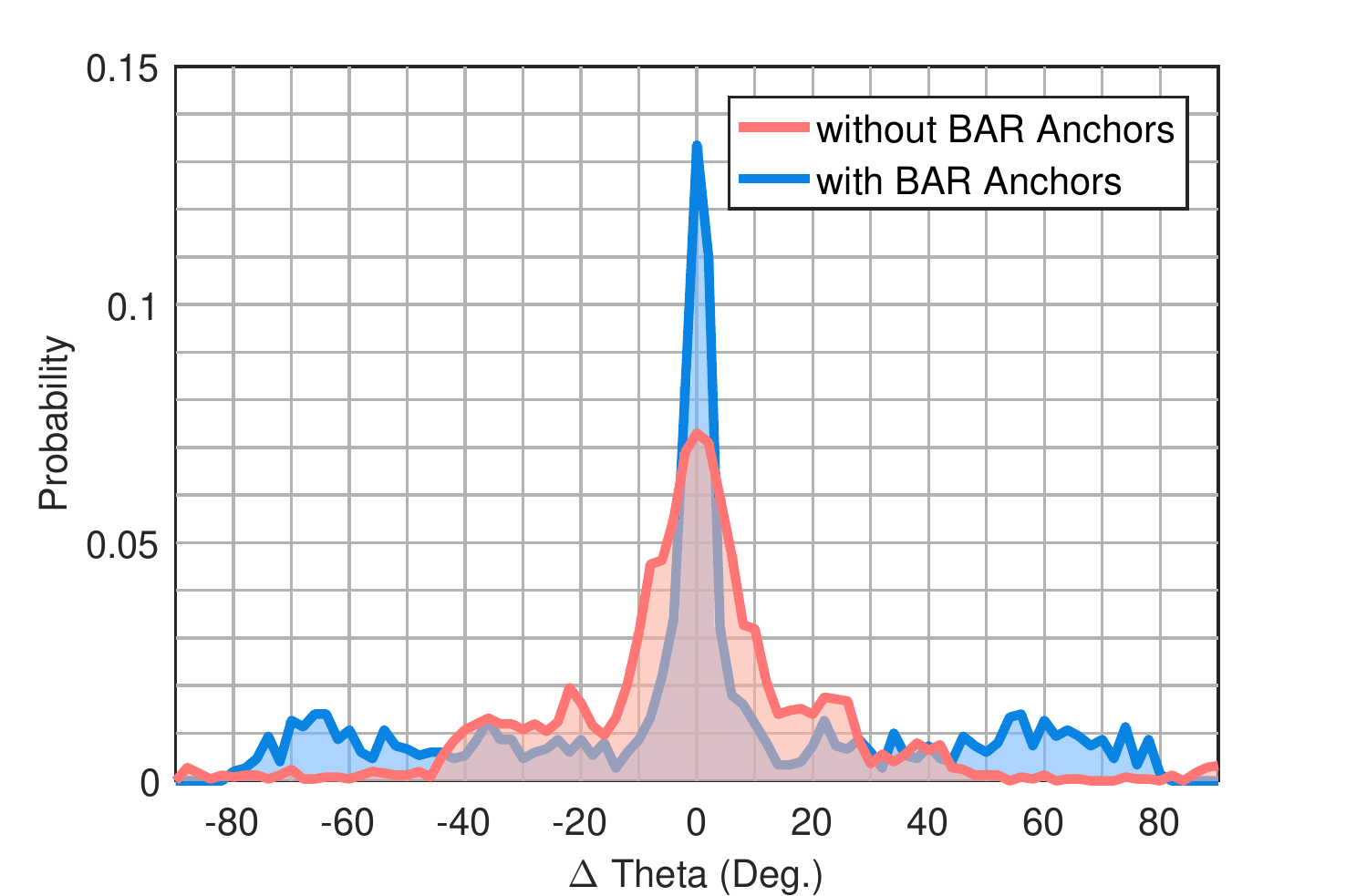}
\caption{The statistic of deviations of various vehicle angles on DLR 3K Munich Dataset.}
\label{fig:theta}
\end{figure}

\textbf{Orientation Accuracy Analysis.} In addition to evaluating localization accuracy, for multi-oriented vehicle detection tasks, we also need to assess the performance of the proposed R$^3$-Net in terms of vehicle orientation estimation. Thus, we make a statistic to show the probability of the deviation $\Delta \theta$ ($-90^{\circ} < \Delta \theta \leq 90^{\circ}$) between predicted angles and ground truth angles. In Fig.~\ref{fig:theta}, we compare two anchor generation strategies, i.e., using the proposed BAR anchor and traditional anchor~\cite{RenShaoqing2015FasterRCNN}, for our method (using ResNet-101 with FPN as feature extraction network) on DLR 3K Munich Dataset. It can be seen that applying BAR anchor to region proposal network can offer better estimations of vehicle orientations, which may attribute to the prior information of vehicle sizes for anchored areas.

\textbf{The Analysis of the Number of Proposals.} As one of important parameters of two-stage detection frameworks, the number of proposals always influences the trade-off between detection accuracy and processing time. However, with the increase of the number of proposals, the detection accuracy can not get continuous increase. Here, we test the recall rate with different numbers of R-RPN proposals and R-DN proposals in the proposed R$^3$-Net (with ResNet-101) on DLR 3K Munich Dataset and VEDAI Dateset, respectively. We set the IoU value to 0.5. In Table~\ref{table:Recall-proposal}, we display the recall rates with different proposal number settings. We set the number of proposals in R-RPN and that in R-DN in the range of $\{100,300,500,1000,2000\}$ and $\{300,500,1000,2000,3000\}$, respectively. It can been seen that for both DLR 3K Munich Dataset and VEDAI Dataset, when the number of proposals in R-RPN is more than 500 and that in R-DN is set more than 1000, respectively, the recall rate is nearly the same. We, therefore, use 500 and 1000 as the abovementioned numbers for our following experiments.

\textbf{Loss Weight Analysis.} When training our network with the joint loss, there are several important hype-parameters (i.e., $\eta$, $\lambda_1$, and $\lambda_2$) which control weights of all components of the loss function. We, therefore, conduct a series of experiments to seek an optimal combination of them. In our method, the loss weight $\eta$ is to balance the weight of R-RPN and R-DN, and $\lambda_1$ and $\lambda_2$ are to balance the weight of classification tasks and regression tasks in R-RPN and R-DN, respectively. We first fix $\lambda_1$ and $\lambda_2$ ($\lambda_1 = \lambda_2 = 1$) and tweak $\eta$, and then assess the influence of $\lambda_1$ and $\lambda_2$ by fixing $\eta$. In our experiments, we use recall rate under IoU $= 0.5$ to see the performance of different parameter settings. In Table~\ref{table:Recall-eta}, we show the trend of recall rate in relation to $\eta$. It can be seen that $\eta=1$ is a turning point of recall rates. I.e., when $\eta$ is smaller than 1, the recall rate increases, and it decreases when $\eta$ is larger than 1. Hence, we set $\eta=1$ for a good trade-off between R-RPN loss and R-DN loss.

\begin{figure*}[tb]
\centering
\includegraphics[width=180mm]{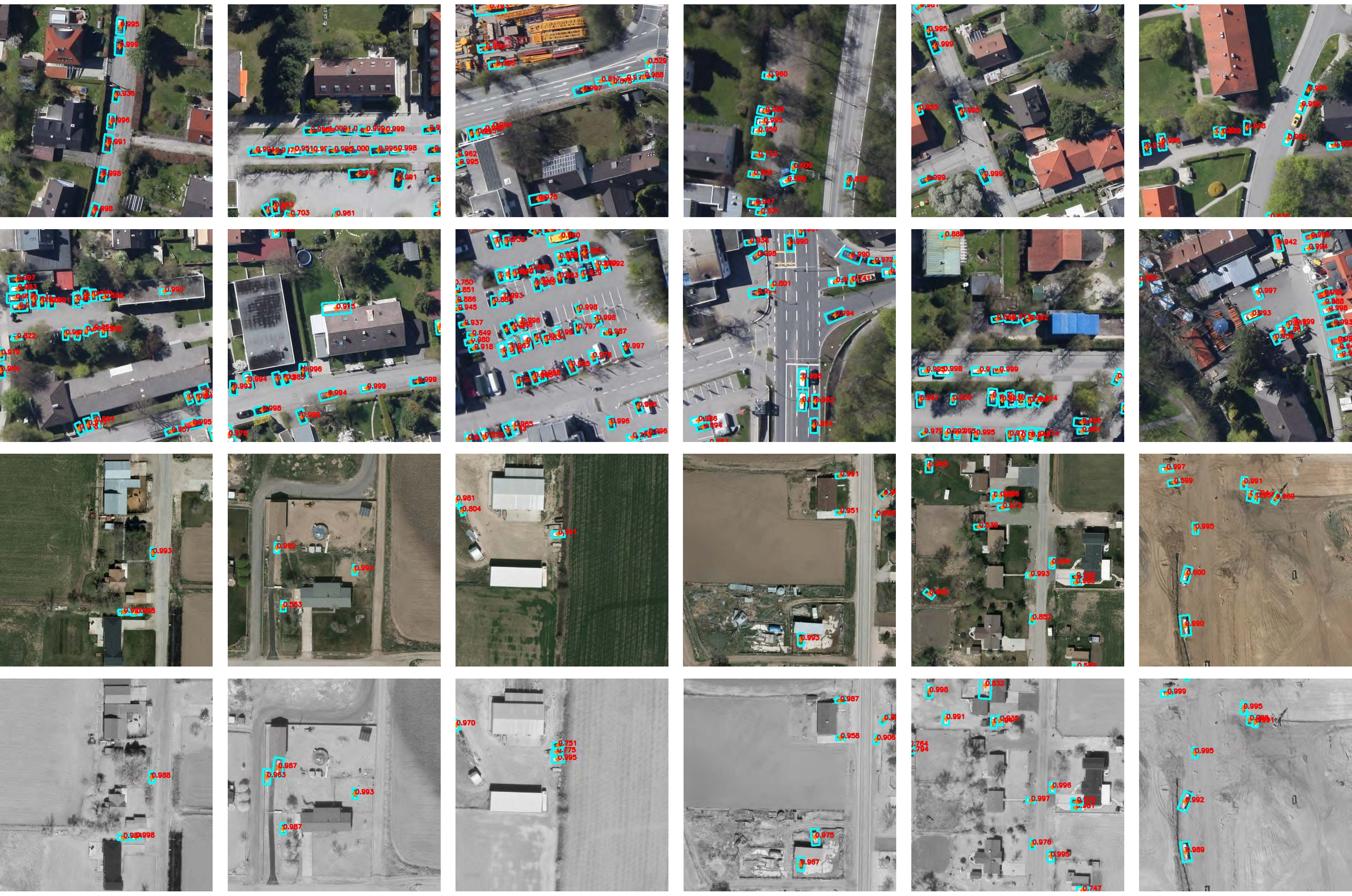}
\caption{Example multi-oriented vehicle detections of the proposed method. First two rows: test samples in DLR 3K Munich Dataset. Last two rows: test samples of VEDAI Dataset (two image modes: RGB and NIR). Best viewed zoomed in.}
\label{fig:samples}
\end{figure*}

Besides, it is necessary to conduct experiments to find out the balance between $\lambda_1$ and $\lambda_2$, which can be tweaked against \emph{gradient domination}. On the one hand, we set $\lambda_2 = 10$ and observe $\lambda_1$ in a range from 0.01 to 100 with seven values. According to the recall rates on DLR 3K Munich Dataset and VEDAI Dataset, we choose $\lambda_1 = 1$ for DLR 3K Munich Dataset and $\lambda_1 = 10$ for VEDAI Dataset. On the other hand, we set $\lambda_1 = 1$ and $\lambda_1 = 10$ for DLR 3K Munich Dataset and VEDAI Dataset, respectively, and then select $\lambda_2$ in the range of 0.01 to 100 with seven values. At last, we find that $\lambda_2 = 10$ is optimal for both of two datasets. From Table~\ref{table:Recall-eta}, we can see that there is no \emph{gradient domination}, which can prove that the proposed method is robust.

\subsection{Comparisons on detection task with other methods}
We compare the proposed network (based on VGG-16, ResNet-101, and ResNet-101 with FPN) with two one-stage CNN-based object detection methods (i.e., SSD\footnote{\url{https://github.com/weiliu89/caffe/tree/ssd}}~\cite{2015SSD} based on VGG-16 and YOLO\footnote{\url{https://github.com/pjreddie/darknet}}~\cite{2015YOLO} based on VGG-16), a two-stage CNN-based object detection method Faster R-CNN \footnote{\url{https://github.com/rbgirshick/py-faster-rcnn}}~\cite{RenShaoqing2015FasterRCNN} based on VGG-16, and a baseline method R-FCN\footnote{\url{https://github.com/YuwenXiong/py-R-FCN}}~\cite{Dai2016RFCN} (based on VGG-16, ResNet-101, and ResNet-101 with FPN).

\begin{table}[t]
\centering
\caption{Recalls of Different Numbers of R-RPN and R-DN Proposals. (R$^3$-Net with ResNet-101, IoU = 0.5)}
\begin{tabular}{l  ccccc}
\toprule[1pt]
\toprule[1pt]
Num. of R-RPN proposals                           & 100    & 300      & 500   & 1000   & 2000  \\
Num. of R-DN proposals                            & 300    & 500      & 1000  & 2000   & 3000 \\
\toprule[1pt]
DLR 3K Munich Dataset                             & 0.605  & 0.753    & 0.809& 0.812  & 0.816 \\
VEDAI Dataset                                     & 0.426  & 0.522    & 0.586& 0.589  & 0.592 \\
\toprule[1pt]
\toprule[1pt]
\end{tabular}
\label{table:Recall-proposal}
\end{table}

\begin{table}[t]
\centering
\caption{Recalls with Different Loss Weights. (R$^3$-Net with ResNet-101, IoU = 0.5)}
\begin{tabular}{l  ccccc}
\toprule[1pt]
\toprule[1pt]
$\eta$ ($\lambda_1=\lambda_2=1$)              & 0.01   & 0.1     & 1   & 10    & 100 \\
\toprule[1pt]
DLR 3K Munich Dataset                         & 0.730  & 0.764   & 0.798& 0.792  & 0.753 \\
VEDAI Dataset                                 & 0.508  & 0.520   & 0.565& 0.537  & 0.528 \\
\toprule[1pt]
\toprule[1pt]
$\lambda_1$ ($\eta=1$, $\lambda_2=10$)         & 0.01   & 0.1      & 1     & 10    & 100 \\
\toprule[1pt]
DLR 3K Munich Dataset                          & 0.741  & 0.766    & 0.809 & 0.781  & 0.769 \\
VEDAI Dataset                                  & 0.536  & 0.558    & 0.571 & 0.586  & 0.562 \\
\toprule[1pt]
\toprule[1pt]
$\lambda_2$ ($\eta=1$, $\lambda_1=\{1,10\}$)   & 0.01   & 0.1      & 1    & 10    & 100 \\
\toprule[1pt]
DLR 3K Munich Dataset                          & 0.733  & 0.778    & 0.798& 0.809  & 0.790 \\
VEDAI Dataset                                  & 0.528  & 0.545    & 0.569& 0.586  & 0.575 \\
\toprule[1pt]
\toprule[1pt]
\end{tabular}
\label{table:Recall-eta}
\end{table}

\textbf{Parameter Settings.} Before further comparison of our method with others, we set proper parameters for each method. Moreover, the Faster R-CNN is implemented based on the Caffe framework, and we generate 2000 proposals by selective search algorithm~\cite{Uijlings2013SelectiveSearch}. Other parameter settings of those CNN-based methods can refer to the open source code.

\begin{figure*}[t]
\centering
\subfigure{
\includegraphics[width=88mm]{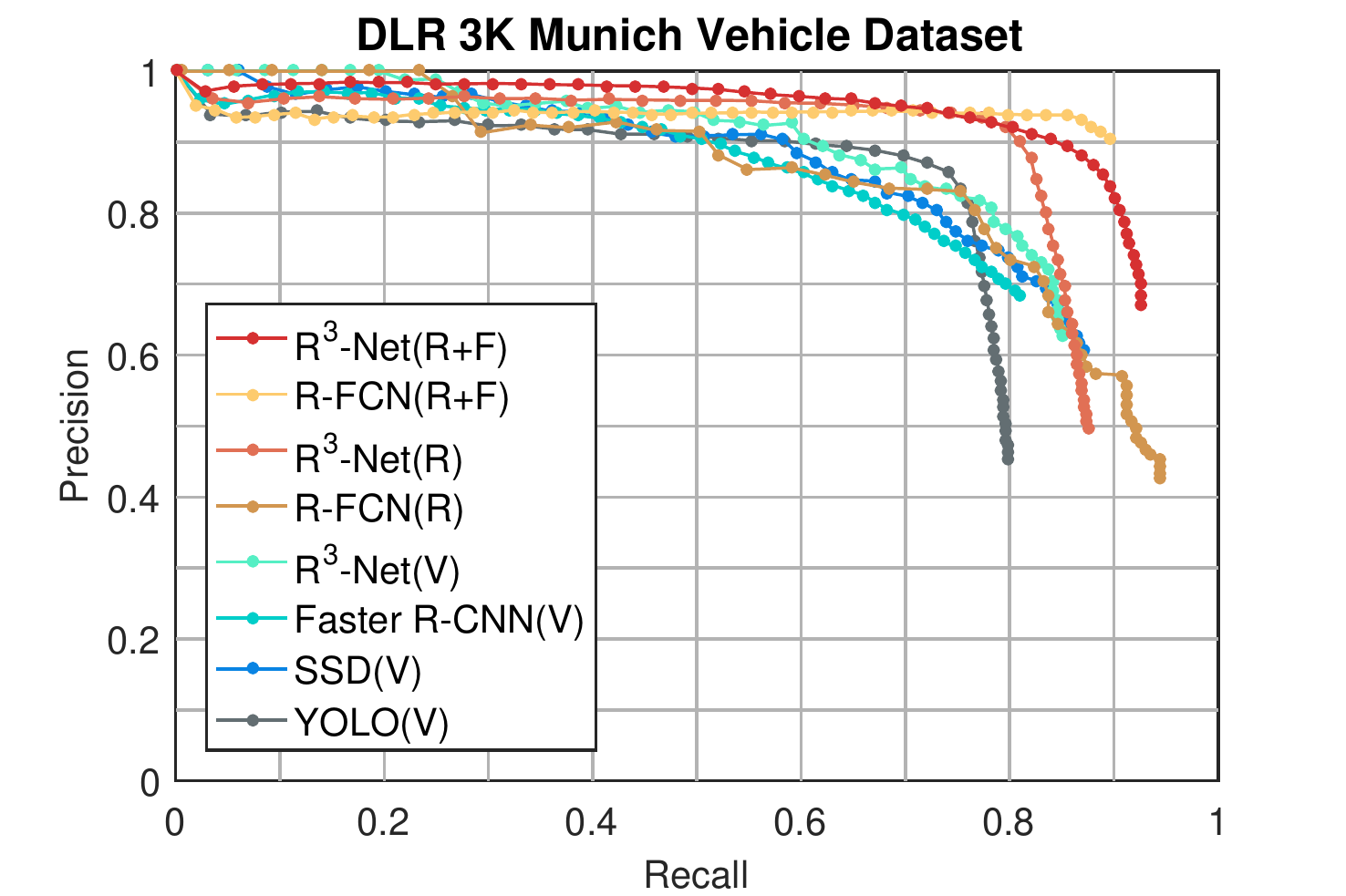}
\includegraphics[width=88mm]{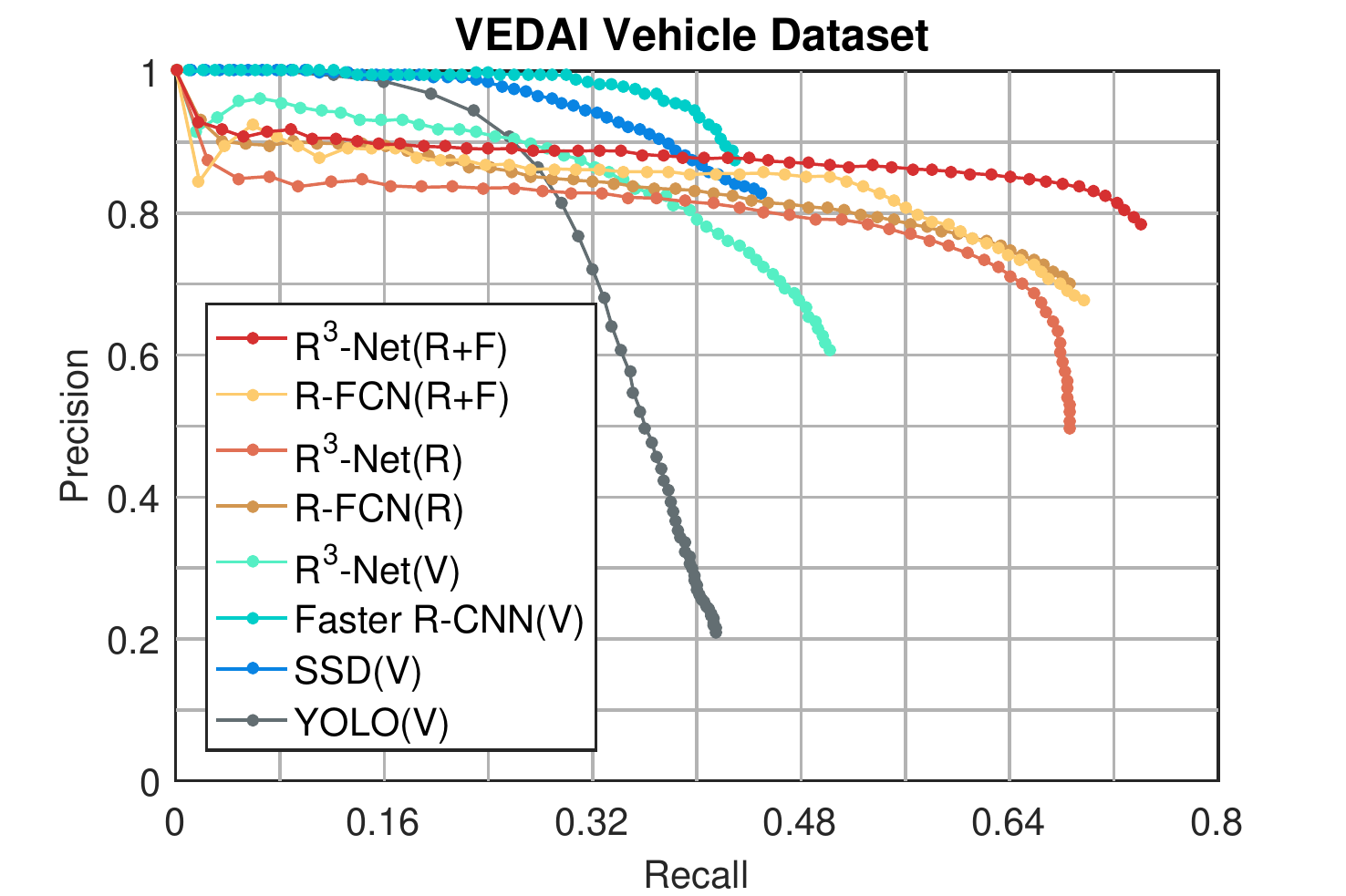}}
\caption{Precision-recall comparisons between R$^3$-Net and other methods on DLR 3K Munich Dataset (left) and VEDAI Dataset (right). (${\rm IoU}=0.6$). Key: (V) = VGG-16; (R) = ResNet-101; (R+F) = ResNet-101 with FPN.}
\label{fig:pr} %% label for entire figure
\end{figure*}

\textbf{Precision-Recall Curve and AP.} We show precision-recall curves and APs of our method and other competitors on both DLR 3K Munich Dataset and VEDAI Dataset, respectively, in Fig.~\ref{fig:pr} and Table~\ref{table:AP}.

On DLR 3K Munich Dataset and VEDAI Dataset, we set the recall threshold from 0 to 1 and 0 to 0.8, respectively, and show Precision-Recall curves of different methods. It can be seen that two-stage CNN-based detection methods are with higher accuracies than one-stage ones, which can be attributed to the fact that two-stage classifiers can obtain more accuracy shots for classification tasks than one-stage ones.

On the other hand, from Table~\ref{table:AP}, in comparison with the baseline model, we can see that the proposed method fails to raise the AP performance too much and even drop down the AP when taking VGG-16 and ResNet-101 as feature extraction networks. However, the proposed method can promote the AP performance a lot by using features extracted by ResNet-101 together with FPN, which probably thanks to the efficient feature fusion mechanism in FPN, contributing to more precise localization for outputting bounding boxes in the shape of rotatable rectangles. Besides, it can be seen that deeper networks are able to offer better results with their stronger capabilities of nonlinear representation.

\begin{table}[t]
\centering
\caption{Average Precision(\%) ($ {\rm IoU} = 0.6 $). Key: (V) = VGG-16; (R) = ResNet-101; (R+F) = ResNet-101 with FPN.}
\begin{tabular}{p{2.9cm} cc}
\toprule[1pt]
\toprule[1pt]
Method & DLR 3K Munich Dataset & VEDAI Dataset     \\
\toprule[1pt]
YOLO(V)                         &71.4\% &36.7\% \\
SSD(V)                          &74.7\% &43.8\% \\
Faster R-CNN(V)                 &73.4\% &44.8\% \\
R$^3$-Net(V)                    &74.2\% &45.2\% \\
R-FCN(R)                        &80.1\% &53.2\% \\
R$^3$-Net(R)                    &79.5\%  &53.4\%  \\
R-FCN(R+F)                      & 85.9\% & 61.8\%  \\
R$^3$-Net(R+F)                  & \textbf{87.0\%}  & \textbf{64.8\%}   \\
\toprule[1pt]
\toprule[1pt]
\end{tabular}
\label{table:AP}
\end{table}

\begin{table}[t]
\centering
\caption{Average Precision(\%) ($ {\rm IoU} = 0.6 $). Key: (V) = VGG-16; (R) = ResNet-101; (R+F) = ResNet-101 with FPN.}
\begin{tabular}{l cc}
\toprule[1pt]
\toprule[1pt]
Method &  DLR 3K Mun. (F1)  & 4-cls VEDAI (mAP)      \\
\toprule[1pt]
Viola-Jones$^{*}$~\cite{2015MunichBaseline}  &0.61 & $-$ \\
Liu's$^{*}$~\cite{2015MunichBaseline}        &0.77 & $-$ \\
AVPN\_basic$^{*}$~\cite{2017AVPN}            & 0.80             & $-$ \\
AVPN\_large$^{*}$~\cite{2017AVPN}            & 0.82             & $-$ \\
Fast R-CNN(AVPN)$^{*}$~\cite{2017AVPN}       & 0.82             & $-$ \\
\toprule[1pt]
DPM$^{*}$~\cite{2015VEDAIBaseline}           & $-$             & 0.46 \\
SVM+LTP$^{*}$~\cite{2015VEDAIBaseline}       & $-$             & 0.51 \\
SVM+HOG31+LBP$^{*}$~\cite{2015VEDAIBaseline} & $-$             & 0.50 \\
\toprule[1pt]
R$^3$-Net(V), IoU = 0.5                   &0.83             &0.47 \\
R$^3$-Net(R), IoU = 0.5                   &0.85             &0.56  \\
R$^3$-Net(R+F), IoU = 0.5                 & \textbf{0.91}   & \textbf{0.69}   \\
\toprule[1pt]
\toprule[1pt]
\end{tabular}
\label{table:index}
\end{table}

In Table~\ref{table:index}, we report F1-scores of several methods, including Viola-Jones detector~\cite{2001VJ}, Liu's method~\cite{2015MunichBaseline}, and AVPN method with different settings~\cite{2017AVPN}. Moreover, we also report mean APs (mAPs) of Deformable Part-based Model (DPM)~\cite{2010DPM} and some traditional detectors which use hand-crafted features (e.g., LBP~\cite{2002LBP}, HOG~\cite{2005HOG}, and LTP~\cite{2010LTP}) on four selected vehicle classes, i.e., car, pickup, truck, and van. The results show that the proposed method outperforms others.

\begin{figure*}[t!]
\centering
\subfigure{
\includegraphics[width=180mm]{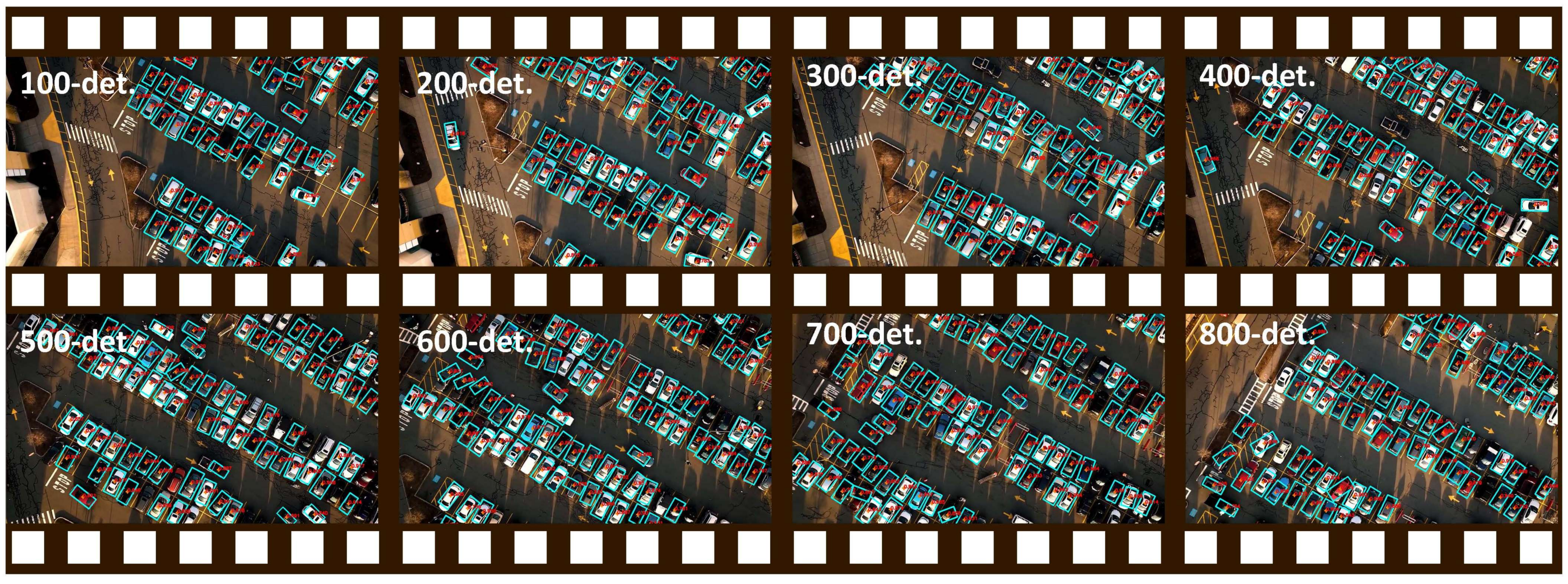}}
\subfigure{
\includegraphics[width=180mm]{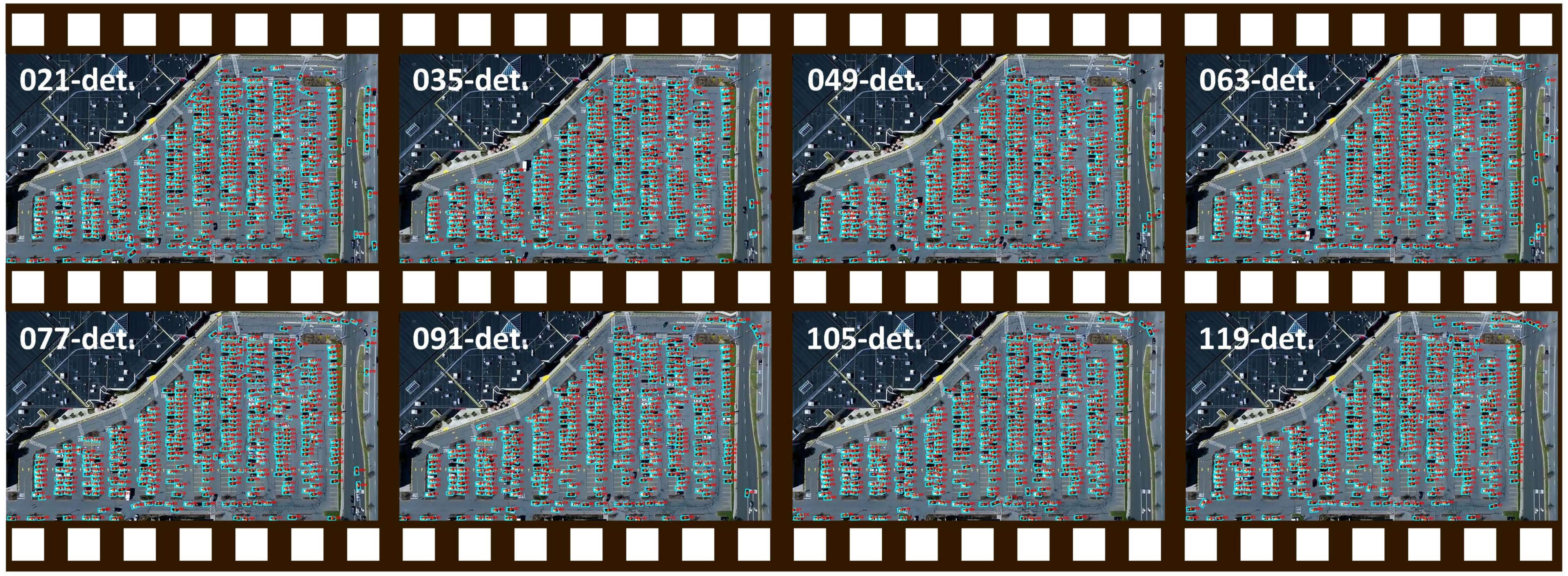}}
\subfigure{
\includegraphics[width=180mm]{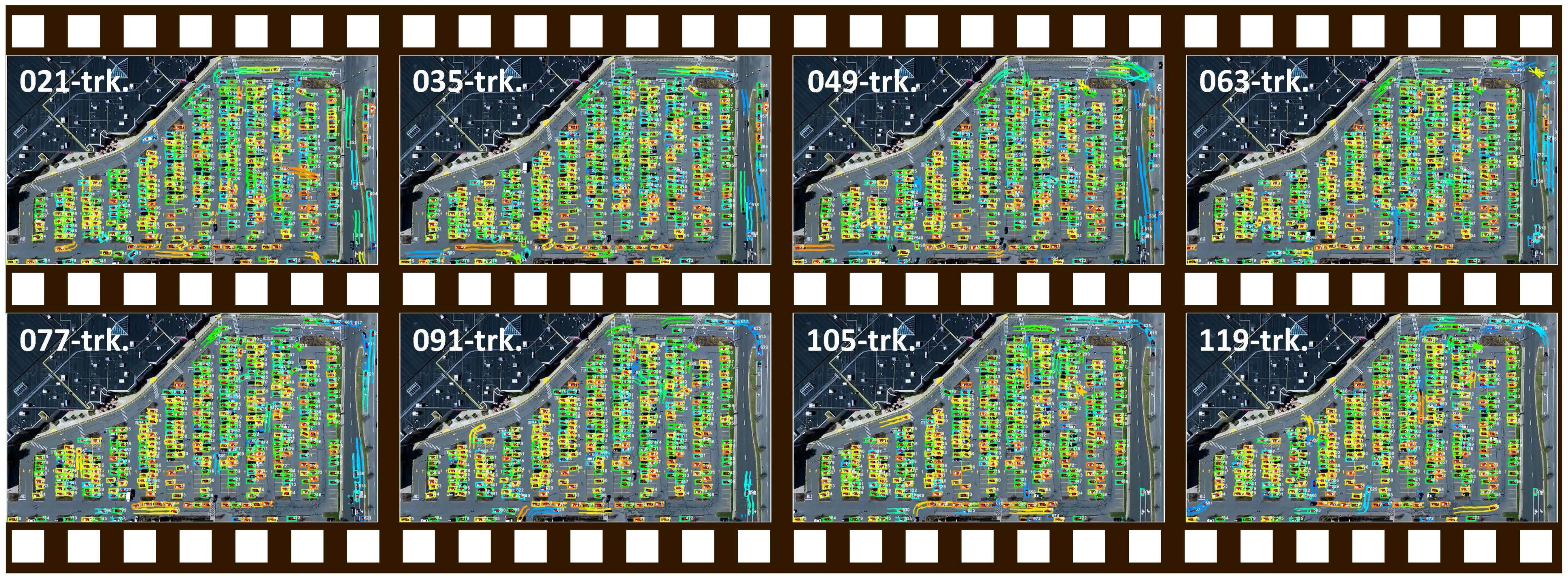}}
\caption{Detection and tracking results on two video datasets. The first 2 rows show the detection result on UAV Parking Lot UAV Cruise Video; The first 2 rows show the detection result on Busy Parking Lot UAV Surveillance Video; The first 2 rows show the detection result on Busy Parking Lot UAV Surveillance Video. Best viewed zoomed in. A part of tracking result is available at \url{https://youtu.be/xCYD-tYudN0}. Key: Det. = Detection; Trk. = Tracking.}
\label{fig:video}
\end{figure*}

\begin{table*}[t!]
\caption{Detection Results on Two UAV Videos. On UAV Video 1 (Cruise Mode), we utilize R$^3$-Net (with ResNet-101 and FPN) as detection method. On UAV Video 2 (Surveillance Mode), we utilize R$^3$-Net (with ResNet-101 and FPN) as detection method together with Kalman filter as tracking method. Key: Det. = Detection; Trk. = Tracking.}
\centering
\begin{tabular}{lcccc cccc cccc}
\toprule[1pt]
\toprule[1pt]
      {Task}&{Mode}&{Frame size}&{Frames}&a-FPS&a-TPs&a-FNs&a-FPs&Precision&Recall&Num. of proposals&Test speed\\
\toprule[1pt]
      {Det.}  &  Cruise        &  $1280\times720$        &  1080           & 24.0                & {67}    & {6}    & {0}   & {100.0\%} & {93.8\%}    &1000 / 500        &10.3 fps     \\
      {Det.}  &  Surveillance         &  $1920\times1080$       &  1452           & 24.2                & {435}   & {46}   & {2}   & {99.3\%}  & {90.4\%}    &5000 / 3000       &1.8 fps     \\
      {Det. by trk.$^\dag$}   &  Surveillance         &  $1920\times1080$       &  1452           & 24.2            & {439}   & {42}   & {0}   & {100.0\%} & {91.3\%}    &5000 / 3000 &1.6 fps    \\
\toprule[1pt]
\toprule[1pt]
\end{tabular}
\label{table:video}
\end{table*}
\begin{table*}[t!]
\caption{Comparison with Instance Segmentation-Based Detection Methods on Four Labeled Frames on UAV Video 2. Key: (R) = ResNet-101; (R+F) = ResNet-101 with FPN.}
\centering
\begin{tabular}{l cccc cccc cccc cccc}
\toprule[1pt]
\toprule[1pt]
\multirow{2}{*}{Method}  & \multicolumn{4}{c} {Frame@1s} & \multicolumn{4}{c} {Frame@15s} & \multicolumn{4}{c} {Frame@30s} & \multicolumn{4}{c} {Frame@45s} \\
\cmidrule[1pt]{2-17}
&{F$_1$}&{Pre.}&{Rec.}&{Tim.} &{F$_1$}&{Pre.}&{Rec.}&{Tim.} &{F$_1$}&{Pre.}&{Rec.}&{Tim.} &{F$_1$}&{Pre.}&{Rec.}&{Tim.} \\
\toprule[1pt]
B-Xception-FCN$^*$~\cite{2018MouSeg}& 91.4 & 89.7 &  \textbf{93.2} & 2.59 & 90.2 & 86.8 & \textbf{93.8} & 2.76  & 90.1 & 87.7 &  92.7          & 2.81 &90.4 & 87.6 & 93.2 & 2.74    \\
B-ResFCN$^*$~\cite{2018MouSeg}& 93.3 & 95.2 &  91.5 & 1.54 & 92.6 & 91.5 & 93.6 & 1.67  & 93.6 & 94.0 &  \textbf{93.2} & 1.92 &93.1 & 94.3 & 91.8 & 1.77    \\
Mask R-CNN(R+F)        & 95.3 & \textbf{99.8} &  91.2            & \textbf{0.24} & 95.6 & 99.6 & 91.9 & \textbf{0.25}  & 95.0 & 99.6 &  90.8 & \textbf{0.25} &95.0 & 99.3 & 91.1   & \textbf{0.24}    \\
R$^3$-Net(R+F)         & 94.9 & 99.6 &  90.6            & 0.56 & 94.9 & 99.3 & 90.9                       & 0.55  & 94.3 & 99.3 &  89.8                   & 0.56 &95.1 & 99.8 & 90.9   & 0.55    \\
R$^3$-Net(R+F)+KF$^\dag$& \textbf{95.5} & \textbf{99.8} &  91.6 & 0.63 & \textbf{95.9} & \textbf{100.0} & 92.3 & 0.63  & \textbf{95.8} & \textbf{99.8}&92.0 & 0.63 &\textbf{96.5} & \textbf{99.8} & \textbf{93.4}& 0.62    \\
\toprule[1pt]
\midrule[1pt]
\end{tabular}
\label{table:seg}
\end{table*}

\subsection{Experiments on aerial videos}
In addition to image data, we also test our method on two aerial videos (see Fig.~\ref{fig:video}), Parking Lot UAV Cruise Video and Busy Parking Lot UAV Surveillance Video~\cite{2018MouSeg}. The former is captured in low altitude cruise mode, and the latter is acquired by a camera onborad a UAV hovering above the parking lot of Woburn Mall, Woburn, MA, USA\footnote{\url{https://www.youtube.com/watch?v=yojapmOkIfg}}. We apply our model (R$^3$-Net with ResNet-101 and FPN trained on DLR 3K Munich Vehicle Dataset) to Parking Lot UAV Cruise Video for detection task to assess the performance of the proposed model on video frames in a dynamic scenery and on Busy Parking Lot UAV Surveillance Video for detection as well as tracking to test the detector's performance in a crowed and complex scene from a relatively static surveillance view. Furthermore, we are curious to know if the temporal information of the second video data can strengthen our detection network's performance. The detection results are shown in Table~\ref{table:video}.

\textbf{Detection on Parking Lot UAV Cruise Video.} In order to qualitatively evaluate our model on this video, we manually labeled 20 ground truths for 20 frames and then compute average TPs, average FNs, average FPs, precision, and recall based on these ground truths. For simplicity, we use transferred regular rectangle boxes to compute IoU as mentioned before. The IoU threshold is set to 0.5. Table~\ref{table:video} shows that the precision is 100\%, and the recall is 93.8\%, which indicates that our trained model has a high \emph{generalization ability} on this video. Besides, the test speed with a frame size of $1280\times720$ is about 10.3 fps, which can nearly satisfy the requirement of a real-time vehicle detection task using UAV videos with a low speed cruise mode. Here, the numbers of proposals in R-RPN and R-DN are set to 1000 and 500, respectively.

\textbf{Detection and Tracking on Busy Parking Lot UAV Surveillance Video.} We also test our model on this data with 10 manually labeled frames. In this video, the main challenge for vehicle detection is that a great number of tiny vehicle appear densely. And in Table~\ref{table:video}, we can see that our trained model has a satisfactory flexibility in such case with a precision of 99.3\% and a recall of 90.4\%. Here, we set the numbers of proposals in R-RPN and R-DN  to 5000 and 3000, respectively, and the speed during test phase with a frame size of $1920\times1080$ is 1.8 fps, which can nearly meet the requirement of a high-altitude surveillance platform.

To facilitate our research, we try to perform multiple object tracking task on Busy Parking Lot UAV Surveillance Video using the detection results produced by the proposed network. We exploit a simple online and real-time tracking (SORT) algorithm for multiple object tracking in video sequences\footnote{\url{https://github.com/abewley/sort}}, which utilizes a Kalman filter (KF) to predict tracking boxes. From the result shown in Table~\ref{table:video}, it can be seen that the capability of the proposed detection network can be upgraded with a precision of 100\% and a recall of 91.3\%\footnote{In this paper, we only discuss the result of indicators for detection task rather than tracking task.}, which is attributed to the fact that the relevance of context information in multiple frames can offset the deviation of single-frame detection. The joint speed of detection and tracking during test phase is about 1.6 fps. In Fig.~\ref{fig:video}, we show vehicle IDs, predicted boxes, and vehicle trajectories. A part of the tracking result is available at \url{https://youtu.be/xCYD-tYudN0}.

\begin{figure}[t!]
\centering
\includegraphics[width=88mm]{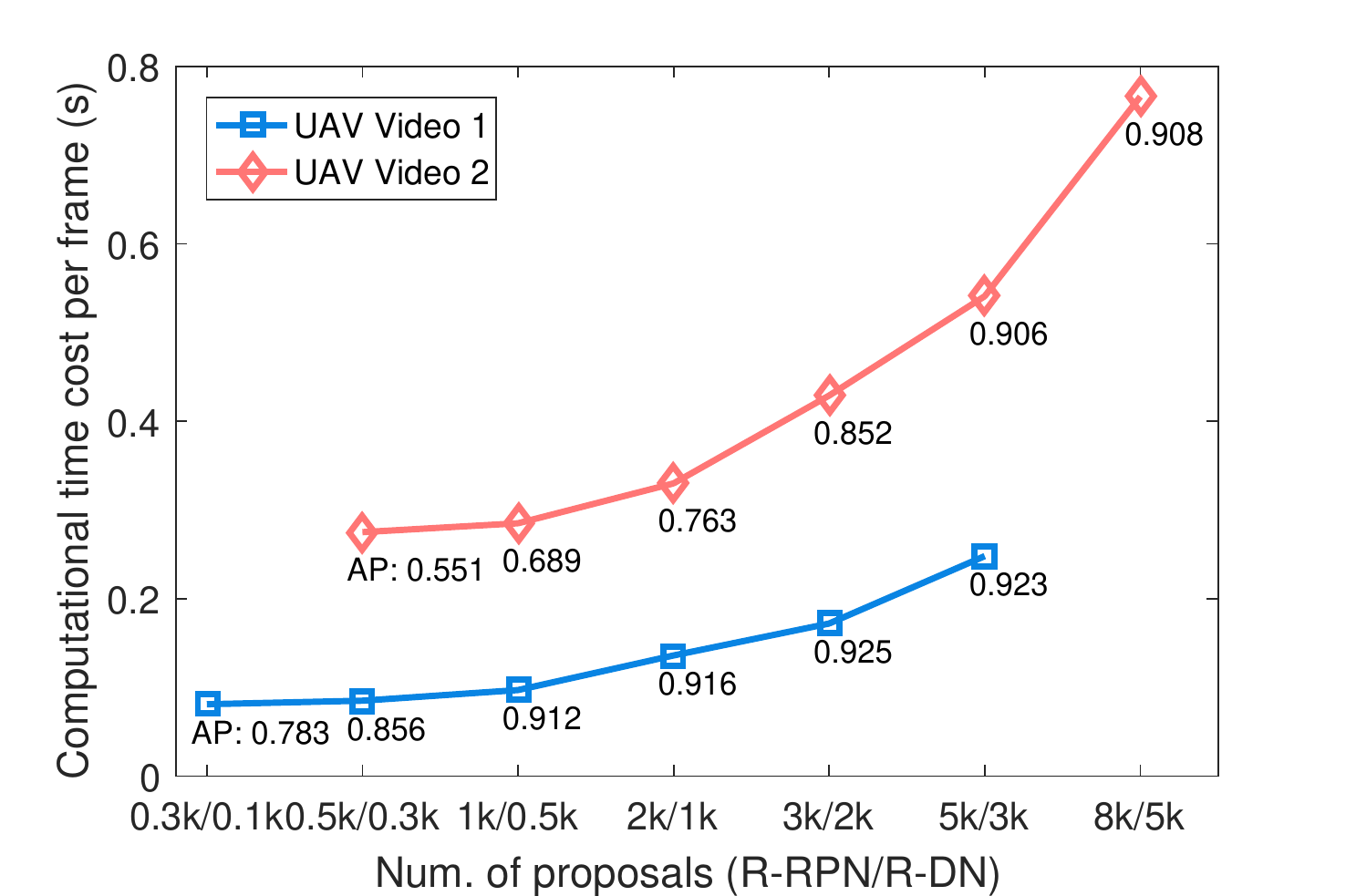}
\caption{Computational time cost per frame in two UAV videos with the change of the number of proposals. Key: AP = Average precision; UAV Video 1 = Parking Lot UAV Cruise Video; UAV Video 2 = Busy Parking Lot UAV Surveillance Video.}
\label{fig:speed}
\end{figure}

\textbf{Comparison with Instance Segmentation-Based Detection Methods.} We compare our method with several state-of-the-art instance segmentation-based detection methods, namely B-Xception-FCN~\cite{2018MouSeg}, B-ResFCN~\cite{2018MouSeg}, and Mask R-CNN~\cite{mask}. Here, B-Xception-FCN model and ResFCN model are trained on ISPRS Potsdam Semantic Labeling Dataset~\cite{2012isprsPotsdam}, and Mask R-CNN model with ResNet-101 and FPN is trained on DLR 3K Munich Dataset. The parameter setting of Mask R-CNN can refer to the open source code\footnote{\url{https://github.com/facebookresearch/Detectron}}. In Table~\ref{table:seg}, we show the comparison on precision, recall, and F$_1$-score. We find that the proposed models get lower recall than those instance segmentation-based methods in general, however, they have better performance on precision and F$_1$-score, which shows the satisfactory flexibility of the proposed method on this video data.

\textbf{Trade-off between Accuracy and Test Time.} In Fig.~\ref{fig:speed}, we evaluate the test time cost and average precision of our method with different numbers of proposals on the two labeled UAV video data in order to find a good trade-off between the accuracy and time cost. As a result, we set the numbers of proposals in R-RPN and R-DN to 1000 and 500 for Parking Lot UAV Cruise Video, and 5000 and 3000 for Busy Parking Lot UAV Surveillance Video, respectively.

\section{Conclusions}
\label{sec:conclusion}
In this paper, a novel method is proposed to detect multi-oriented vehicles in aerial images and videos using a deep network call R$^3$-Net. First, one typical CNN is utilized to extract deep features. Second, we use R-RPN to generate R-RoIs encoded in 8-d vectors. A novel strategy called BAR anchor is applied to initialize templates of rotatable candidates. Third, we use R-DN as classifier and regressor to obtain the final 5-d rotatable detection boxes. Here, we propose a new downsampling method for R-RoIs called R-PS pooling to achieve fast dimensionality reduction on R-RoI feature maps and keep the information of positions and orientations. Besides, we modify the Shamos Algorithm for the conversion of 5-d and 8-d detection boxes in R-RPN and R-DN. In our method, R-RPN and R-DN can be jointly trained for high efficiency. Then we evaluate the proposed method from two perspectives. On the one hand, we perform experiments on two open vehicle detection image datasets, i.e., DLR 3K Munich Dataset and VEDAI Dataset, to compare with other state-of-the-art detection methods. On the other hand, we conduct extra experiments on two aerial videos using models trained on DLR 3K Munich Dataset. Experimental results show that the proposed R$^3$-Net outperforms other methods on both aerial images and aerial videos. Especially, R$^3$-Net can be well combined with multiple object tracking methods to acquire further information (e.g., vehicle trajectory), showing the satisfactory performance on multi-oriented vehicle detection tasks.

% Reference
\bibliographystyle{mybibfile}
\bibliography{mybibfile}

\clearpage
\end{document}